\documentclass[pdflatex,sn-mathphys-num]{sn-jnl}

\usepackage{graphicx}%
\usepackage{multirow}%
\usepackage{amsmath,amssymb,amsfonts}%
\usepackage{amsthm}%
\usepackage{mathrsfs}%
\usepackage[title]{appendix}%
\usepackage{xcolor}%
\usepackage{textcomp}%
\usepackage{manyfoot}%
\usepackage{booktabs}%
\usepackage{algorithm}%
\usepackage{algorithmicx}%
\usepackage{algpseudocode}%
\usepackage{listings}%
\usepackage{makecell}
\usepackage{booktabs,tabularx}  
\usepackage{xcolor}            

\theoremstyle{thmstyleone}%
\newcommand\at[2]{\left.#1\right|_{#2}}
\theoremstyle{thmstyletwo}%

\theoremstyle{thmstylethree}%

\begin{document}

\title[Article Title]{Training with Explanations Alone: A New Paradigm to Prevent Shortcut Learning}


\author*[1,2]{\fnm{Pedro R.A.S.} \sur{Bassi}}\email{pedro.salvadorbassi2@unibo.it}

\author[2]{\fnm{Haydr A.H.} \sur{Ali}}

\author[1,2,3]{\fnm{Andrea} \sur{Cavalli}}

\author*[2]{\fnm{Sergio} \sur{Decherchi}}\email{sergio.decherchi@iit.it}

\affil[1]{University of Bologna, Italy}

\affil[2]{Istituto Italiano de Tecnologia, Italy}

\affil[3]{École Polytechnique Fédérale de Lausanne, Swizerland}


\abstract{
Application of Artificial Intelligence (AI) in critical domains, like the medical one, is often hampered by shortcut learning, which hinders AI generalization to diverse hospitals and patients. 
Shortcut learning can be caused, for example, by background biases---features in image backgrounds that are spuriously correlated to classification labels (e.g., words in X-rays). 
To mitigate the influence of image background and foreground bias on AI, we introduce a new training paradigm, dubbed Training with Explanations Alone (TEA). 
TEA trains a classifier (TEA student) only by making its explanation heatmaps match target heatmaps from a larger teacher model. 
By learning from its explanation heatmaps, the TEA student pays attention to the same image features as the teacher.
For example, a teacher uses a large segmenter to remove image backgrounds before classification, thus ignoring background bias. By learning from the teacher's explanation heatmaps, the TEA student learns to also ignore backgrounds---but it does not need a segmenter. 
With different teachers, the TEA student can also resist bias in the image foreground. 
Surprisingly, by training with heatmaps alone the student output naturally matches the teacher output---with no loss function applied to the student output. 
We compared the TEA student against 14 state-of-the-art methods in 5 datasets with strong background or foreground bias, including Waterbirds and an X-Ray dataset for COVID-19 and pneumonia classification. The TEA student had better resistance to bias, strongly surpassing state-of-the-art methods, and generalizing better to hospitals not seen in training.


}

\maketitle
\section{Main}\label{sec:intro}

With deep learning, artificial intelligence (AI) classifiers reached human-like accuracy in multiple computer vision tasks, and are expected to help humans in critical applications including medical ones (e.g. radiology). These applications require trustworthy AI, but a major obstacle for trustworthiness is shortcut learning \cite{geirhos2020shortcut,degrave2021ai}. Shortcut learning is the phenomenon where AI learns decision rules that fail in real-world scenarios, because they rely on bias or spurious correlations found in datasets. For example, during the COVID-19 pandemic, multiple studies reported classifiers with super-human accuracy in classifying COVID-19 in X-rays. However, many of these classifiers would classify an X-ray as COVID-19 not due to the disease symptoms, but due to background bias (text or other features in the background of the X-ray) \cite{degrave2021ai}. Background bias is known to cause shortcut learning \cite{geirhos2020shortcut} and it exists in diverse datasets, including pneumonia and tuberculosis X-ray datasets \cite{bassi2024improving}, and famous computer vision datasets, like PASCAL VOC \cite{lapuschkin2019unmasking}. Therefore, a key question for reducing shortcut learning and advancing towards trustworthy AI is: \textit{how can we train classifiers to pay attention to meaningful image features rather than background bias?}

To this end, different approaches were proposed. The most straightforward one is a segmentation-classification pipeline: a segmenter finds and removes the image background, and a classifier classifies the segmented image. E.g., segment the lungs before classifying lung diseases in X-rays \cite{bassi2022covid}. However, this classifier depends on the segmentation step to be accurate even at inference. As the segmenter is often large, the segmentation-classification pipeline can be slow and difficult to deploy in low-resource settings, like embedded devices or medical institutions in developing regions \cite{bassi2024improving}. To avoid this higher deployment cost, studies proposed training standard classifiers to avoid attention to bias, through background attention minimization in explanation heatmaps (e.g., ISNet \cite{bassi2024improving}, GAIN \cite{li2018tell}, and RRR \cite{ross2017right}). These studies use explanation techniques (e.g., Grad-CAM \cite{selvaraju2017grad}, input gradients \cite{simonyan2014deep}, and Layer-wise relevance propagation/LRP \cite{bach2015onpixel}) to create explanation heatmaps that represent the attention of the classifier during training. Then, a heatmap-based loss---used alongside the classification loss---minimizes the attention in the backgrounds of the  explanation heatmaps, teaching the classifiers to disregard the background. However, the heatmap-based loss function and the classification loss function are opposing forces. Attention in the background increases the heatmap-based loss function, but decreases the classification loss function when background bias is present. In applications where the background bias is too extensive, the opposing loss functions may prevent the convergence of the classifier to a solution that has high accuracy and resistance to bias (see Tab. \ref{tab:coco_Waterbirds_results}).

\begin{figure}
    \centering
    \includegraphics[width=1\linewidth]{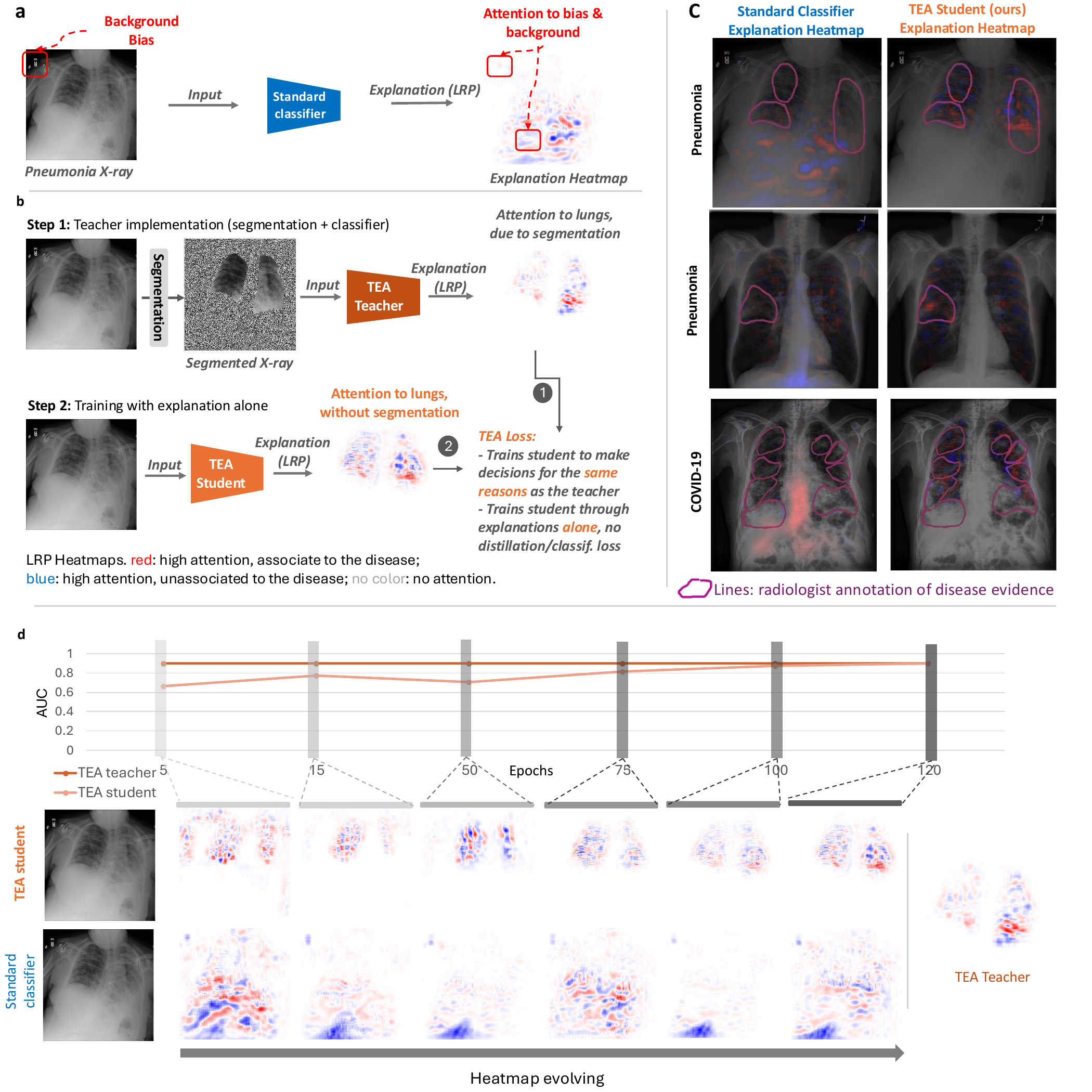}
    \caption{\textbf{The Training with Explanations Alone paradigm.} \textbf{(a)} \textbf{Baseline: a standard classifier trained to classify pneumonia and COVID-19 in our X-ray dataset has strong background attention.} Attention to background bias (letters in X-ray corner) and irrelevant background regions (outside of lungs) is evident (red and blue in LRP heatmaps). \textbf{(b)} \textbf{TEA methodology.} A frozen, heavy teacher (e.g., a pipeline that segments lungs, then classifies lung diseases) classifies the images, and we generate its explanation heatmaps. The lighter TEA student (no segmenter) classifies the image, and we generate its explanation heatmaps. The TEA loss trains the student, minimizing the dissimilarity between the its heatmaps and teacher's. Through heatmaps, we teach the student the reasons behind the teacher's classification outputs. As the teacher ignores background bias, the student also ignores it. To better avoid bias attention, we train the student though explanation heatmaps alone---no classification or distillation loss is applied to its outputs or intermediate representations. \textbf{(c)} \textbf{The correlation between where the TEA student and a radiologist found COVID-19 and pneumonia evidence is stronger than the correlation between the standard classifier and the radiologist.} Radiologist annotations are plum lines, and areas where classifiers found COVID-19 or pneumonia evidence are red (LRP heatmaps). Radiologists had no access to heatmaps. \textbf{(d)} \textbf{In training, the TEA student AUC gradually approximates the teacher's, as the TEA student heatmaps approximate the teacher's.} Thus, by learning the reasons behind the teacher outputs (expressed in heatmaps), the TEA student outputs and AUC gradually become similar to the teacher's, with no loss function applied to the student outputs.}
    \label{fig:method}
\end{figure}

Here, we propose a training paradigm that solves the problem of opposing loss functions and is resistant to background bias. Specifically, we propose abandoning the classification loss function (and any loss function applied to the classifier's outputs or intermediate representations\footnote{We use "outputs" to refer to the outputs of the last layer of the classifier, and "intermediate representations" to refer to the outputs of any  previous layer.}).
Instead of training by matching classifier outputs to target class labels, we train only by matching classifier explanation heatmaps to target heatmaps---associated to the right outputs and with no attention to background bias. Explanation heatmaps created by techniques like LRP can explain how each pixel in the input image contributed to the classifier output.
By optimizing the explanation heatmaps of a classifier to match the target explanation heatmaps, the classifier learns to produce right outputs based on the right image features---not bias. 
We named this paradigm ``\textbf{Training with Explanations Alone}'' or TEA. Here, we obtain target explanation heatmaps (without background bias attention) from a segmentation-classification pipeline (i.e., the ``teacher''), and we use them to train a standard classifier (i.e., the ``TEA student''). The TEA student is faster and lighter than the segmentation-classification pipeline (teacher), by not requiring a segmenter. Most importantly, it is more accurate and resistant to extensive background biases than background attention minimization techniques like RRR and ISNet, because it is not trained with opposing loss functions. TEA also surpasses other teacher-student techniques, including standard output distillation \cite{hinton2015distilling}, advanced intermediate representation distillation \cite{lee2023debiased,vapnik2009learning}, and even the distillation of explanation heatmaps plus outputs \cite{parchami2024good}. 
The long-standing teacher-student paradigm directly requires the student to match the output or intermediate representations of the teacher, but the student can leverage bias and shortcut learning to achieve this goal (Tab. \ref{tab:ablations}). Conversely, TEA only requires the student to learn the reasons behind the teacher outputs---expressed in explanation heatmaps. By learning these reasons, the student inherits the teacher's resistance to bias, and its outputs (and accuracy) naturally become similar to the teacher's.
This is the first work to show it is possible to train classifiers with explanations alone. From the cognitive viewpoint, this paradigm resembles the real interaction between a human student and teacher, where the teacher explains the reasons behind answers. 
TEA can use different techniques to create explanation heatmaps. We experimented with five techniques, with LRP coming on top (Tab. \ref{tab:ablations}). LRP back-propagates a signal through the entire classifier, capturing high-level information from late layers, and precise spatial information from early ones. Furthermore, LRP provides a stronger theoretical support for TEA (Sec. \ref{sec:explanations}). Fig. \ref{fig:method} summarizes the TEA methodology. 

We present extensive experiments to assess the TEA student resistance to bias, using 5 datasets and comparing it to 14 state-of-the-art methods. First, we experiment with two well known datasets of photographs with extensive background bias, COCO-on-Places \cite{ahmed2020systematic} and Waterbirds \cite{sagawa2019distributionally}. The biases in these datasets are synthetic (objects or birds are copied and pasted over different background scenes, like lakes). Thus, backgrounds can be changed in the test set, allowing us to quantitatively measure their influence on classifiers (Sec. \ref{sec:background_results_synth}). The extensive background bias strongly influenced state-of-the-art background attention minimization techniques, but it had minimal influence on the TEA student. In both datasets, the TEA student surpassed all 13 tested state-of-the-art methods in resistance to bias. These methods include background attention minimization (ISNet, GAIN, and RRR), distillation techniques \cite{hinton2015distilling,lee2023debiased,parchami2024good}, and group robustness (IRM \cite{arjovsky2019invariant}, PGI \cite{ahmed2020systematic}, GroupDRO \cite{sagawa2019distributionally}).


Next, we experimented on the COVID-19 and pneumonia X-ray dataset from \cite{bassi2024improving}. Background biases exist in diverse popular X-ray datasets, but it is less extensive than the synthetic biases in COCO-on-Places and Waterbirds (covering a smaller portion of the image) \cite{degrave2021ai,bassi2024improving,bassi2022covid}. Our X-ray training set \cite{bassi2024improving} was designed to be similar to most COVID-19 X-ray dataset during the first years of the pandemic, where X-rays for different classes (COVID-19, pneumonia or normal) originated from different hospitals (mixed-source dataset). X-rays from different hospitals present different background features (e.g., text), and these features become spuriously correlated to the classes in mixed-source dataset---becoming background bias \cite{bassi2024improving,degrave2021ai}. For example, many early COVID-19 X-rays were from Italy, while most control X-rays were not. Thus, some classifiers learned to consider Italian text in the X-ray background as an evidence of COVID-19 \cite{bassi2022deep}. Since background bias is known to hinder classifier generalization to unseen hospitals, we use an out-of-distribution (OOD) test set \cite{bassi2024improving}, with X-rays from hospitals not seen in the training set. By minimizing attention to features in the background (outside of the lungs), TEA mitigated shortcut learning and improved performance in the unseen test hospitals (Tab. \ref{tab:x_ray_results}). 

Although this study mainly focuses on background bias, our last experiments show TEA can also be extended to resist biases in the foreground\footnote{In this study, ``foreground bias'' means colors or objects in the foreground of the image.}, unlike background attention minimization techniques. To demonstrate this, we use two datasets: Coloured MNIST \cite{arjovsky2019invariant}, a digit classification dataset where each digit is spuriously correlated to a color in the training data; and our new DogsWithTies dataset, a dog breed classification dataset where each breed is associated with a different tie shape and color. The TEA student resisted to biases in the foreground and significantly surpassed all tested state-of-the-art methods. 


We provide code to easily implement TEA for any ReLU-based classifier architecture in PyTorch, available at \url{https://github.com/PedroRASB/TrainingWithExplanationsAlone}. By this code, researchers can easily train new architectures with TEA, making it ignore biases without any additional computational cost at inference time. In summary, the contributions of this work are:
\begin{enumerate}
    \item We show deep classifiers can be trained by optimizing their explanation heatmaps alone, instead of their outputs or intermediate representations. A student classifier, whose explanation heatmaps are optimized to match a teacher's explanation heatmaps, learns to produce classification outputs similar to the teacher---without ever having access to the teacher outputs (Tabs. \ref{tab:coco_Waterbirds_results}, \ref{tab:x_ray_results}, \ref{tab:foreground}).
    \item We show that training with explanations alone yields more resistance to extensive background bias, in comparison to multiple state-of-the-art methods and to training with explanations plus standard distillation loss functions (Tabs. \ref{tab:coco_Waterbirds_results}, \ref{tab:ablations}).
    \item We show training with explanations alone can also result in strong resistance to biases in the foreground (Tab. \ref{tab:foreground}).
\end{enumerate}



\section{Results}

\begin{figure}[t]
    \centering
    \includegraphics[width=1\linewidth]{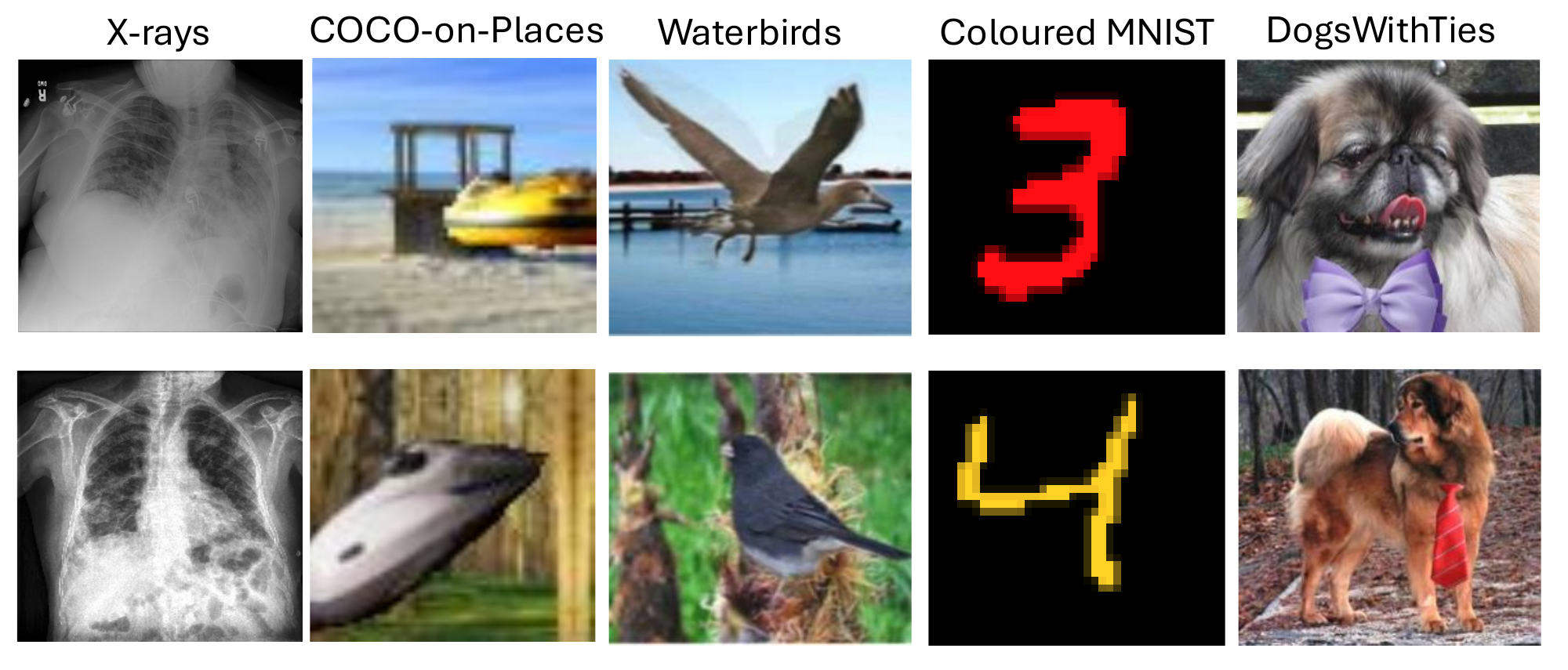}
    \caption{\textbf{Examples of images from each training dataset.} Bias is notable. E.g., biases in the top row are: a marking in the top left corner of the X-ray (pneumonia X-ray, background bias), the beach background behind the boat (COCO-on-Places, background bias), the water background behind the waterbird (Waterbirds, background bias), the red color in the digit 3 (Coloured MNIST, foreground bias), and the purple bow tie in the Pekingese dog (DogsWithTies, foreground bias).}
    \label{fig:dataset_examples}
\end{figure}

\begin{figure}[t]
    \centering
    \includegraphics[width=1\linewidth]{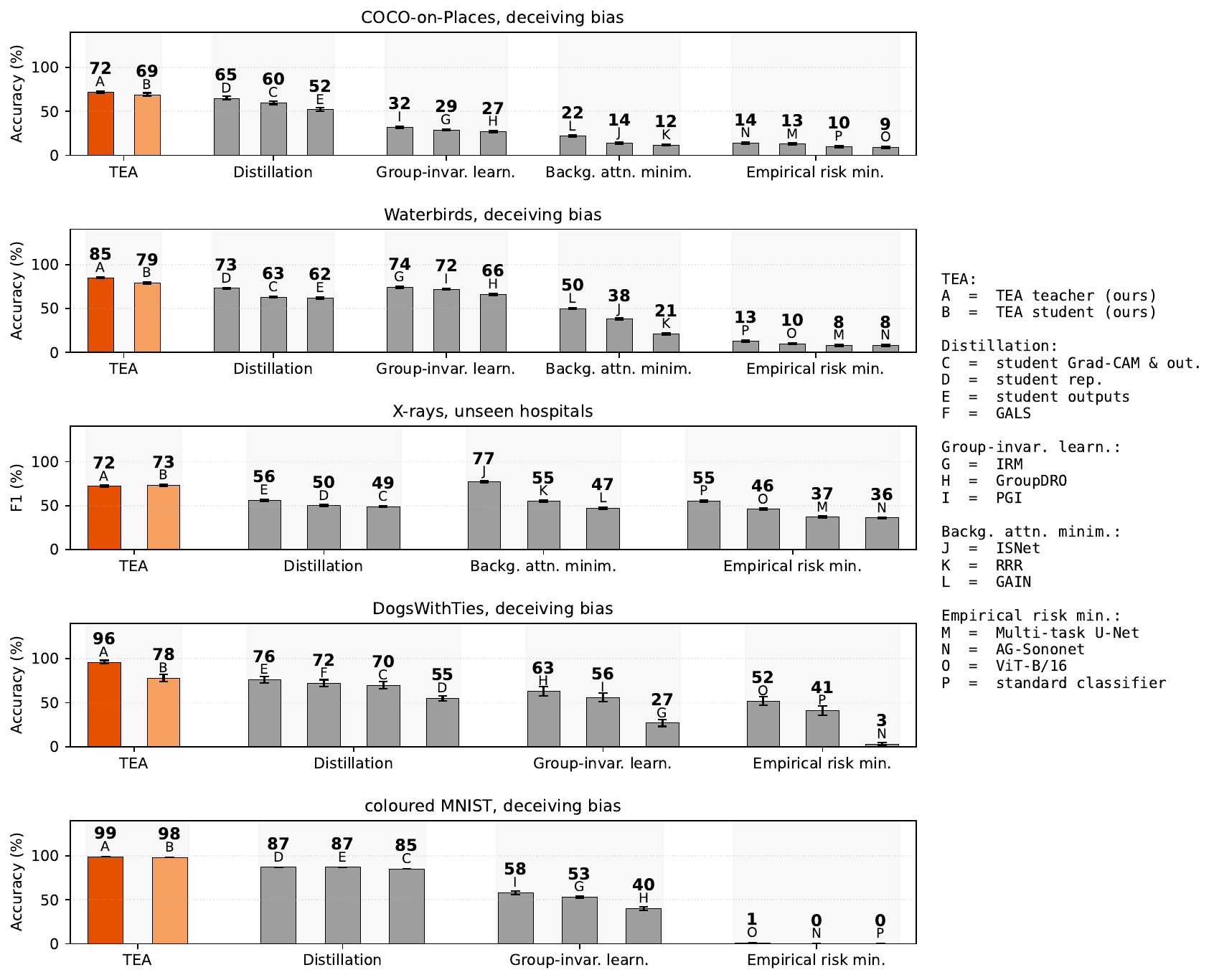}
    \caption{\textbf{TEA surpasses state-of-the-art methods in resistance to bias.} The TEA student surpasses all methods on 4 of 5 datasets. It is only surpassed by 1 method (ISNet \cite{bassi2024improving}) in 1 dateset (X-rays)---because the ISNet surpasses even the TEA teacher in this dataset. However, the TEA student strongly surpasses the ISNet in COCO-on-Places and Waterbirds, being more resistant to the more extensive background bias there. Four plots show balanced accuracy in deceiving bias test sets: Coloured MNIST and DogsWithTies (foreground bias), Waterbirds and COCO-on-Places (background bias). Deceiving bias is set to fool bias-reliant classifiers, and accuracy on deceiving bias sets is also known as worst-group accuracy. E.g., in training, all waterbirds are over water and landbirds over land; in deceiving bias, all waterbirds are over land and landbirds over water. To perform well with deceiving bias, classifiers resist bias. For X-rays, we test on hospitals never seen during training and plot average F1-Score. As X-ray bias is not synthetic, there is no deceiving bias test set. However, better performance in unseen hospitals is linked to better bias resistance \cite{degrave2021ai}. Error bars represent standard deviation (Supp. Note \ref{app:statistical}). Test set sizes are: N=2,897 in Waterbirds, N=900 in COCO-on-Places, N=3,180 in X-rays, N=10,000 in Coloured MNIST, and N=101 in DogsWithTies.}
    \label{fig:results}
\end{figure}

Figure \ref{fig:results} summarizes the TEA strong resistance to bias in all 5 datasets, which are summarized in Tab. \ref{tab:datasets} and exemplified in Fig. \ref{fig:dataset_examples}. These datasets have different nature, biases, sizes, and resolutions: COCO-on-Places and Waterbirds (Sec. \ref{sec:background_results_synth}) have natural images (photographs) with extensive synthetic background bias, large size (N=9K in COCOonPlaces, N=12K in Waterbirds) and medium resolution (64p); the X-ray dataset (Sec. \ref{sec:background_results_x_ray}) has medical images, non-synthetic background bias, large size (N=17K) and resolution (224p); Coloured MNIST (Sec. \ref{sec:foreground_results}) has digits with synthetic foreground bias (color), large size (N=60K) and low resolution (28p); and DogsWithTies has dog photographs with synthetic foreground bias (ties), small size (N=261) and high resolution (224p). Sections \ref{sec:background_results_synth} to \ref{sec:foreground_results} detail results on each dataset. Sec. \ref{sec:ablations_results} presents ablation studies on explanation techniques and loss functions used, and demonstrates that training with explanations alone (TEA) creates stronger resistance to bias than training with explanations plus outputs/intermediate representation distillation. 

\begin{table}[t]
\centering
\caption{\textbf{Summary of datasets.} To thoroughly test TEA, we use 5 datasets with diverse types of image, sizes, resolutions, types of bias, and bias location. The objective is classifying the classes, while ignoring the bias. In COCO-on-Places classes are objects, and biases are the background scenes. In Waterbirds, classes are bird types (waterbird/landbird), and biases are the land or water backgrounds. In the X-ray dataset, classes are normal, COVID-19 and non-COVID-19 Pneumonia, and biases are markings and features outside of the lungs. This mixed-source dataset intentionally mirrors early COVID-19 data: each class comes from different hospitals, so hospital-specific characteristics in X-ray backgrounds correlate with class labels and create background bias. In Coloured MNIST, classes are hand written digits, and biases are their colors. In DogsWithTies classes are dog breeds (Tibetan Mastiff/Pekingese) and biases are ties on their necks.}
\label{tab:datasets}
\footnotesize
\setlength{\tabcolsep}{4pt}
\begin{tabular*}{\linewidth}{@{\extracolsep{\fill}} lccccccc}
\toprule
\textbf{Dataset} & \textbf{Train N} & \textbf{Val N} & \textbf{Test N} & \textbf{Res.} & \textbf{Image type} & \textbf{Bias} & \textbf{Bias Loc.} \\
\midrule
COCO-on-Places~\cite{ahmed2020systematic}       & 7,200 & 900   & 900    & 64p   & photographs       & synthetic   & background \\
Waterbirds~\cite{sagawa2019distributionally}    & 4,315 & 480   & 2,897 & 64p   & photographs       & synthetic   & background \\
\makecell[l]{COVID-19/Pneumonia\\X-rays (mixed-source)~\cite{bassi2024improving}} & 10,449 & 3,483 & 3,216 & 224p & \makecell{medical (X-ray)}           & real  & background \\
Coloured MNIST~\cite{arjovsky2019invariant}     & 54,000 & 6,000 & 10,000 & 28p   & \makecell{handwritten\\digits} & synthetic   & foreground \\
DogsWithTies                                    & 180   & 20    & 101    & 224p & photographs       & synthetic   & foreground \\
\bottomrule
\end{tabular*}
\end{table}

Training \& data processing / augmentation details are in Supp. Note \ref{app:training_augmentation}, dataset details in \ref{app:dataset_details}, and state-of-the-art methods details in Supp. Note \ref{app:related}. Many of these state-of-the-art methods are designed to avoid bias attention---background attention minimization and group robustness methods. Distillation is also shown to avoid bias attention \cite{lee2023debiased}. ViT and AG-Sononet have attention mechanisms but are \textit{not originally intended} to avoid bias attention. TEA and most of the tested state-of-the-art methods are training paradigms applicable to many classifier architectures\footnote{Three state-of-the-art methods are more intrinsically related to architectures: the multi-task U-Net \cite{MultiTask2}, AG-Sononet \cite{schlemper2018attention} and Vision Transformer (ViT-B/16) \cite{dosovitskiy2020image}. In these cases, we use the original architectures proposed in the respective papers in all of our experiments.}. Here, we use the ResNet-50 \cite{he2016deep} for COCO-on-Places and Waterbirds, due to its popularity; DenseNet121 for X-rays, due to its high accuracy in X-rays \cite{rajpurkar2017chexnet}; and ResNet-18 for Coloured MNIST and DogsWithTies, to demonstrate a small TEA student can learn from a large teacher (ResNet-50 and CLIP). Our X-ray experiments use a dataset designed to assess classifiers resistance to bias \cite{bassi2024improving}. It mimics early COVID-19 datasets, which commonly had background bias. See dataset limitations in Supp. Note \ref{app:dataset_details}.


\subsection{TEA resists extensive background bias in photographs}
\label{sec:background_results_synth}










Table \ref{tab:coco_Waterbirds_results} shows the TEA student resistance to background bias surpassed 13 state-of-the-art methods in both the COCO-on-Places and Waterbirds datasets. The datasets' tasks consist in classifying 10 object classes in COCO-on-Places, and 2 birds classes in Waterbirds (waterbirds, bird species that live around water Vs. landbirds). In both datasets, background bias is extensive: the entire background is synthetically modified to be spuriously associated to these classes. E.g., waterbirds are placed over water and landbirds over land in all Waterbirds training images\footnote{We can choose the percentage of training images with background bias in Waterbirds and COCO-on-Places. We use the hardest setting for the TEA student: 100\% of biased images (Waterbirds-100\% and COCO-on-Places-100\%).}. We quantitatively measured resistance to background bias through a generalization gap: $|\mathrm{Acc}_{b} - \mathrm{Acc_{db}}|$. $\mathrm{Acc}_{b}$ is the classification accuracy on a biased test dataset, which has the same background biases as the training data (e.g., waterbirds always appear over water in the training and biased testing test datasets). $\mathrm{Acc_{db}}$ is the classification accuracy on a deceiving bias test set, where correlations between biases and classes were switched to deceive classifiers (e.g., waterbirds always appear over water in training and over land in the deceiving bias test dataset). The goal is to obtain a low generalization gap and high accuracy in all test sets. Conversely, high accuracy only in the test biased set reveals biased classifiers. In Tab. \ref{tab:coco_Waterbirds_results}, a standard classifier had massive generalization gaps (64$\pm$1\% in COCO-on-Places, 82$\pm$1\% in Waterbirds), showing that the extensive bias in these datasets can strongly influence classifiers' decisions. 

The TEA student is resistant to bias, with the smallest (p$<$0.05) generalization gaps in both datasets: 4$\pm$1\% in COCO-on-Places and 2$\pm$1\% in Waterbirds. These gaps are 2 to 43 times smaller than the 13 state-of-the-art methods we tested, and they are only surpassed by the TEA teacher. Importantly, the teacher needs a foreground segmenter even at inference---without it, accuracy can drop by up to 60\%. The TEA student does not need a segmenter at inference, making it 108\% faster and 5$\times$ more compact than the teacher (Supp. Note \ref{app:speed})---a key quality for deployment in embedded devices and medical institutions in developing regions.


\begin{threeparttable}[t]
\centering
\caption{\textbf{The extensive background bias in the COCO-on-Places and Waterbirds datasets has minimal influence on the TEA student, which surpasses all 13 state-of-the-art methods we tested.} The table sows balanced accuracy and generalization gaps. Small gaps indicate better resistance to bias. The TEA student gap was smaller than all state-of-the-art methods, almost rivaling its teacher, a segmentation-classification pipeline 5 times larger and 2 times slower than the TEA student. All models were trained on datasets with bias covering the entire image background (e.g., waterbirds over water backgrounds, and landbirds over land). To measure resistance to bias, we use different test sets. \textbf{(I)} Biased test set: all test images have background biases like in training (e.g., waterbirds over water). \textbf{(II)} Unbiased test set: no background bias, backgrounds are unseen places (available in COCO-on-Places only).  \textbf{(III)} Deceiving bias test set: the correlation between background biases and classes is switched to deceive classifiers (e.g., waterbirds are over land). The accuracy difference across the three test sets is small for classifiers resistant to background bias. Generalization gap indicates the difference in accuracy between the biased and deceiving test sets (the smaller, the better). Balanced accuracy is shown as mean +/-std (see Supp. Note \ref{app:statistical} for details on the statistical analysis). For group robustness, 80\% of the training images had background bias in COCO-on-Places (harder dataset, 10 classes), and 95\% in Waterbirds. For all other models, including TEA, 100\% of the training images had background bias---a very difficult setting. Image resolution is 64p, making the datasets more difficult (lower accuracy) than 224p versions, but making training faster. Tab. \ref{tab:x_ray_results} shows experiments on a 224p dataset, and Tab. \ref{tab:foreground} on 64p and 28p datasets. Test set sizes are N=2,897 in Waterbirds, and N=900 in COCO-on-Places. Except for those in the architectures section, all classifiers are ResNet50.}
\label{tab:coco_Waterbirds_results}
\scriptsize
\begin{tabular}{%
  p{0.3\textwidth}                               
  *{4}{>{\centering\arraybackslash}p{0.09\textwidth}} 
  *{3}{>{\centering\arraybackslash}p{0.09\textwidth}} 
}
\toprule
 & \multicolumn{4}{c}{COCO-on-Places} &
   \multicolumn{3}{c}{Waterbirds} \\
\cmidrule(lr){2-5}\cmidrule(lr){6-8}
method &
biased\ acc & unbiased\ acc & deceiv. bias\ acc & gen.\ gap $\downarrow$ &
biased\ acc & deceiv. bias\ acc & gen.\ gap $\downarrow$ \\
\midrule
\multicolumn{8}{l}{\textit{empirical risk minimization}}\\
Multi‑task U‑Net \cite{MultiTask2} & 80\textsubscript{$\pm$1} & 25\textsubscript{$\pm$1} & 13\textsubscript{$\pm$1} & 67\textsubscript{$\pm$1} & 94\textsubscript{$\pm1$} & 8\textsubscript{$\pm1$} & 86\textsubscript{$\pm1$} \\
AG‑Sononet \cite{schlemper2018attention}            & 81\textsubscript{$\pm$1} & 27\textsubscript{$\pm$1} & 14\textsubscript{$\pm$1} & 68\textsubscript{$\pm$1} & 94\textsubscript{$\pm1$} & 8\textsubscript{$\pm1$} & 86\textsubscript{$\pm1$} \\
ViT‑B/16 \cite{dosovitskiy2020image}              & 64\textsubscript{$\pm$1} & 15\textsubscript{$\pm$1} & 9\textsubscript{$\pm$1} & 54\textsubscript{$\pm$1} & 93\textsubscript{$\pm1$} & 10\textsubscript{$\pm1$} & 83\textsubscript{$\pm1$} \\
standard classifier \cite{he2016deep} & 73\textsubscript{$\pm$1} & 21\textsubscript{$\pm$1} &  10\textsubscript{$\pm$1} & 64\textsubscript{$\pm$1} & 94\textsubscript{$\pm1$} & 13\textsubscript{$\pm1$} & 82\textsubscript{$\pm1$} \\
\midrule
\multicolumn{8}{l}{\textit{group robustness}}\\
IRM \cite{arjovsky2019invariant}              & 81\textsubscript{$\pm$1} & 45\textsubscript{$\pm$1} & 29\textsubscript{$\pm$1} & 52\textsubscript{$\pm$1} & 78\textsubscript{$\pm1$} & 74\textsubscript{$\pm1$} & 4\textsubscript{$\pm1$} \\
GroupDRO \cite{sagawa2019distributionally}         & 81\textsubscript{$\pm$0} & 42\textsubscript{$\pm$1} & 27\textsubscript{$\pm$1} & 54\textsubscript{$\pm$1} & 74\textsubscript{$\pm1$} & 66\textsubscript{$\pm1$} & 8\textsubscript{$\pm1$} \\
PGI \cite{ahmed2020systematic}              & 81\textsubscript{$\pm$1} & 48\textsubscript{$\pm$1} & 32\textsubscript{$\pm$1} & 49\textsubscript{$\pm$1} & 80\textsubscript{$\pm1$} & 72\textsubscript{$\pm1$} & 8\textsubscript{$\pm1$} \\
\midrule
\multicolumn{8}{l}{\textit{background attention minimization}}\\
ISNet \cite{bassi2024improving}           & 26\textsubscript{$\pm$1} & 15\textsubscript{$\pm$1} & 14\textsubscript{$\pm$1} & 12\textsubscript{$\pm$1} & 65\textsubscript{$\pm1$} & 38\textsubscript{$\pm1$} & 27\textsubscript{$\pm2$} \\
RRR \cite{ross2017right}             & 29\textsubscript{$\pm$1} & 13\textsubscript{$\pm$1} & 12\textsubscript{$\pm$1} & 18\textsubscript{$\pm$1} & 83\textsubscript{$\pm1$} & 21\textsubscript{$\pm1$} & 62\textsubscript{$\pm1$} \\
GAIN \cite{li2018tell}            & 36\textsubscript{$\pm$1} & 27\textsubscript{$\pm$1} & 22\textsubscript{$\pm$1} & 14\textsubscript{$\pm$1} & 71\textsubscript{$\pm1$} & 50\textsubscript{$\pm1$} & 21\textsubscript{$\pm1$} \\
\midrule
\multicolumn{8}{l}{\textit{distillation}$^\dagger$}\\
TEA teacher$^{\dagger\dagger}$ & 72\textsubscript{$\pm$1} & 72\textsubscript{$\pm$1} & 72\textsubscript{$\pm$1} &  1\textsubscript{$\pm$1} & 85\textsubscript{$\pm1$} & 85\textsubscript{$\pm1$} & 0\textsubscript{$\pm1$} \\
stu. Grad-CAM \& out. \cite{parchami2024good}  & 82\textsubscript{$\pm$1}  & 64\textsubscript{$\pm$2}  & 60\textsubscript{$\pm$2}  & 22\textsubscript{$\pm$2} & 91\textsubscript{$\pm1$} & 63\textsubscript{$\pm1$} & 28\textsubscript{$\pm1$} \\
student represent. \cite{lee2023debiased}$^{\dagger\dagger\dagger}$    & 80\textsubscript{$\pm$1} & 68\textsubscript{$\pm$1} & 65\textsubscript{$\pm$2} & 15\textsubscript{$\pm$2} & 88\textsubscript{$\pm1$} & 73\textsubscript{$\pm1$} & 16\textsubscript{$\pm1$}  \\
student outputs \cite{hinton2015distilling}  & 85\textsubscript{$\pm$1} & 61\textsubscript{$\pm$2} & 52\textsubscript{$\pm$2} & 32\textsubscript{$\pm$2} & 91\textsubscript{$\pm1$} & 62\textsubscript{$\pm1$} & 28\textsubscript{$\pm1$} \\
\midrule
\textbf{TEA student (ours)}  & 74\textsubscript{$\pm$1} & 71\textsubscript{$\pm$2} & 69\textsubscript{$\pm$2} &  4\textsubscript{$\pm$2} & 77\textsubscript{$\pm$1} & 79\textsubscript{$\pm$1} & 2\textsubscript{$\pm$1} \\
\bottomrule
\end{tabular}
\begin{tablenotes}[flushleft]\footnotesize
      \item $\dagger$ For fairness, TEA and all distillation methods use the same teacher (the segmentation–classification pipeline) and the same student architecture (ResNet50 classifier). However, the original papers for the alternative distillation techniques did not propose distilling from a segmentation-classification pipeline.
      \item $\dagger\dagger$ The teacher is a segmentation-classification pipeline, which cannot work without the segmenter, making it slower at inference. Without the segmenter, its accuracy drops to 16\textsubscript{$\pm$1}/14\textsubscript{$\pm$1}/12\textsubscript{$\pm$1} in the COCO-on-Places biased/unbiased/deceiving test sets; and to 79\textsubscript{$\pm$1}/34\textsubscript{$\pm$1} in the Waterbirds biased/deceiving test sets.
      \item $\dagger\dagger\dagger$ Improved distillation of intermediate representations by transplanting and freezing the last layer  \cite{lee2023debiased}.
    \end{tablenotes}
\end{threeparttable}

With the same teacher for every method, TEA strongly surpassed state-of-the-art distillation methods in transferring resistance to bias from the teacher to the student. The TEA student generalization gap was 3.8 to 14 times smaller than the gap for the the students in output distillation, intermediate representation distillation, and distillation of Grad-CAM plus outputs (Tab. \ref{tab:coco_Waterbirds_results}).
TEA compels the student to learn the reasons behind the outputs of the teacher, by training the student with explanation heatmaps that express how each image feature contributed to the teacher outputs. In this learning process, the student outputs naturally become similar to the teacher outputs (both have similar accuracy in Tab. \ref{tab:coco_Waterbirds_results}), and the student learns to ignore bias like the teacher does. Conversely, standard distillation losses compel the student outputs or intermediate representations to be similar to the teacher's, but the reasons for the student outputs/representations can be biased, while the teacher is bias-resistant (Tab. \ref{tab:coco_Waterbirds_results}). Unlike previous work, we avoid standard distillation losses and train the student with explanations alone, yielding higher resistance to background bias (see ablation studies in Tab. \ref{tab:ablations}).

\subsection{TEA resists background bias in X-rays \& improves generalization}
\label{sec:background_results_x_ray}

\begin{table}[!t]
\centering
\caption{\textbf{On an external (unseen) hospital, the TEA student surpassed (p$<$0.05) 10 state-of-the-art methods for detecting COVID-19 and pneumonia from X-rays---due to higher resistance to background bias.} It even matched its teacher (p$<$0.05), a segmentation-classification pipeline 5 times larger and 2 times slower than the student. All models were trained in a mixed-source dataset with background bias \cite{bassi2024improving} outside of the lungs---e.g., text spuriously correlated to the diseases. Higher performance on unseen test hospitals indicates resistance to background bias \cite{bassi2024improving,degrave2021ai}. TEA surpassed all methods except for the ISNet (p$<$0.05). However, this is expected because the ISNet considerably surpasses the TEA teacher (p$<$0.05). The table shows sensitivity (Se), specificity (Sp), F1-Score (F1) and AUC (\%), as mean ± SD. See Supp. Note \ref{app:statistical} for details on statistical analyses. Test set size is N=3,216 (406 normals, 1,295 pneumonia, 1,515 COVID-19). Except for the multi-task U-Net, AG-Sononet and ViT, all classifiers are DenseNet121.}
\label{tab:x_ray_results}

{\scriptsize
\begin{tabular*}{\textwidth}{@{\extracolsep{\fill}} lccccccc}
\toprule
 & \multicolumn{2}{c}{Normal} &
   \multicolumn{2}{c}{Pneumonia} &
   \multicolumn{2}{c}{COVID‑19} &
   Avg \\[2pt]
\cmidrule(lr){2-3}\cmidrule(lr){4-5}\cmidrule(lr){6-7}\cmidrule(lr){8-8}
method & Se & Sp & Se & Sp & Se & Sp & F1 \\
\midrule
standard classifier \cite{huang2017densely}  & 57\textsubscript{$\pm$3} & 87\textsubscript{$\pm$1} & 29\textsubscript{$\pm$1} & 96\textsubscript{$\pm$1} & 92\textsubscript{$\pm$1} & 55\textsubscript{$\pm$1} & 55\textsubscript{$\pm$1} \\
Multi‑task U‑Net \cite{MultiTask2}     & 34\textsubscript{$\pm$2} & 96\textsubscript{$\pm$0} & 8\textsubscript{$\pm$1} & 82\textsubscript{$\pm$1} & 78\textsubscript{$\pm$1} & 20\textsubscript{$\pm$1} & 37\textsubscript{$\pm$1} \\
AG‑Sononet \cite{schlemper2018attention}            & 16\textsubscript{$\pm$2} & 82\textsubscript{$\pm$1} & 18\textsubscript{$\pm$1} & 94\textsubscript{$\pm$1} & 82\textsubscript{$\pm$1} & 38\textsubscript{$\pm$1} & 36\textsubscript{$\pm$1} \\
ViT‑B/16 \cite{dosovitskiy2020image}              & 67\textsubscript{$\pm$2} & 76\textsubscript{$\pm$1} & 42\textsubscript{$\pm$1} & 77\textsubscript{$\pm$1} & 49\textsubscript{$\pm$1} & 67\textsubscript{$\pm$1} & 46\textsubscript{$\pm$1} \\
\midrule
\multicolumn{8}{l}{\textit{background attention minimization}}\\
ISNet \cite{bassi2024improving}                 & 57\textsubscript{$\pm$3} & 94\textsubscript{$\pm$1} & 93\textsubscript{$\pm$1} & 83\textsubscript{$\pm$1} & 84\textsubscript{$\pm$1} & 100\textsubscript{$\pm$0} & 77\textsubscript{$\pm$1} \\
RRR \cite{ross2017right}                    & 57\textsubscript{$\pm$3} & 79\textsubscript{$\pm$1} & 45\textsubscript{$\pm$1} & 89\textsubscript{$\pm$1} & 75\textsubscript{$\pm$1} & 74\textsubscript{$\pm$1} & 55\textsubscript{$\pm$1} \\
GAIN \cite{li2018tell}         & 22\textsubscript{$\pm$2} & 88\textsubscript{$\pm$1} & 41\textsubscript{$\pm$1} & 82\textsubscript{$\pm$1} & 80\textsubscript{$\pm$1} & 60\textsubscript{$\pm$1} & 47\textsubscript{$\pm$1} \\
\midrule
\multicolumn{8}{l}{\textit{distillation}}\\
TEA teacher   & 88\textsubscript{$\pm$2} & 89\textsubscript{$\pm$1} & 59\textsubscript{$\pm$1} & 94\textsubscript{$\pm$1} & 86\textsubscript{$\pm$1} & 78\textsubscript{$\pm$1} & 72\textsubscript{$\pm$1}  \\
stu. Grad-CAM \& out. \cite{parchami2024good} & 48\textsubscript{$\pm$3} & 98\textsubscript{$\pm$0} & 11\textsubscript{$\pm$1} & 99\textsubscript{$\pm$0} & 97\textsubscript{$\pm$0} & 21\textsubscript{$\pm$1} & 49\textsubscript{$\pm$1} \\
student representations \cite{lee2023debiased}   & 66\textsubscript{$\pm$2} & 95\textsubscript{$\pm$0} & 9\textsubscript{$\pm$1} & 99\textsubscript{$\pm$0} & 94\textsubscript{$\pm$1} & 25\textsubscript{$\pm$1} & 50\textsubscript{$\pm$1} \\
student outputs \cite{hinton2015distilling}  & 76\textsubscript{$\pm$2} & 94\textsubscript{$\pm$0} & 18\textsubscript{$\pm$1} & 99\textsubscript{$\pm$0} & 91\textsubscript{$\pm$1} & 34\textsubscript{$\pm$1} & 56\textsubscript{$\pm$1} \\
\midrule
\textbf{TEA student (ours)} & 69\textsubscript{$\pm$2} & 91\textsubscript{$\pm$1} & 70\textsubscript{$\pm$1} & 89\textsubscript{$\pm$1} & 86\textsubscript{$\pm$1} & 85\textsubscript{$\pm$1} & 73\textsubscript{$\pm$1} \\
\midrule
 & \multicolumn{2}{c}{Normal} & \multicolumn{2}{c}{Pneumonia} & \multicolumn{2}{c}{COVID‑19} & Avg \\
\cmidrule(lr){2-3}\cmidrule(lr){4-5}\cmidrule(lr){6-7}\cmidrule(lr){8-8}
method & F1 & AUC & F1 & AUC & F1 & AUC & AUC \\
\midrule
standard classifier \cite{huang2017densely}           & 44\textsubscript{$\pm$2} & 80\textsubscript{$\pm$2} & 43\textsubscript{$\pm$2} & 81\textsubscript{$\pm$2} & 76\textsubscript{$\pm$1} & 86\textsubscript{$\pm$1} & 82\textsubscript{$\pm$2} \\
Multi‑task U‑Net \cite{MultiTask2}      & 42\textsubscript{$\pm$3} & 72\textsubscript{$\pm$3} & 12\textsubscript{$\pm$1} & 41\textsubscript{$\pm$2} & 59\textsubscript{$\pm$1} & 49\textsubscript{$\pm$2} & 54\textsubscript{$\pm$2} \\
AG‑Sononet \cite{schlemper2018attention}            & 12\textsubscript{$\pm$2} & 45\textsubscript{$\pm$3} & 28\textsubscript{$\pm$2} & 68\textsubscript{$\pm$2} & 66\textsubscript{$\pm$1} & 66\textsubscript{$\pm$2} & 60\textsubscript{$\pm$2} \\
ViT‑B/16 \cite{dosovitskiy2020image}              & 38\textsubscript{$\pm$2} & 76\textsubscript{$\pm$3} & 47\textsubscript{$\pm$1} & 65\textsubscript{$\pm$2} & 53\textsubscript{$\pm$1} & 62\textsubscript{$\pm$2} & 67\textsubscript{$\pm$2} \\
\midrule
\multicolumn{8}{l}{\textit{background attention minimization}}\\
ISNet \cite{bassi2024improving}                 & 56\textsubscript{$\pm$2} & 93\textsubscript{$\pm$1} & 86\textsubscript{$\pm$1} & 96\textsubscript{$\pm$1} & 91\textsubscript{$\pm$1} & 98\textsubscript{$\pm$1} & 96\textsubscript{$\pm$1} \\
RRR \cite{ross2017right}                    & 36\textsubscript{$\pm$2} & 78\textsubscript{$\pm$2} & 55\textsubscript{$\pm$1} & 74\textsubscript{$\pm$2} & 74\textsubscript{$\pm$1} & 84\textsubscript{$\pm$1} & 78\textsubscript{$\pm$2} \\
GAIN \cite{li2018tell}         & 20\textsubscript{$\pm$2} & 70\textsubscript{$\pm$3} & 49\textsubscript{$\pm$1} & 76\textsubscript{$\pm$2} & 71\textsubscript{$\pm$1} & 81\textsubscript{$\pm$2} & 75\textsubscript{$\pm$2} \\
\midrule
\multicolumn{8}{l}{\textit{distillation}$^\dagger$}\\
TEA teacher$^{\dagger\dagger}$     & 64\textsubscript{$\pm$2} & 95\textsubscript{$\pm$1} & 70\textsubscript{$\pm$1} & 86\textsubscript{$\pm$1} & 82\textsubscript{$\pm$1} & 90\textsubscript{$\pm$1} & 90\textsubscript{$\pm$1} \\
stu. Grad-CAM \& out. \cite{parchami2024good} & 58\textsubscript{$\pm$2} & 92\textsubscript{$\pm$2} & 20\textsubscript{$\pm$1} & 61\textsubscript{$\pm$2} & 68\textsubscript{$\pm$1} & 70\textsubscript{$\pm$2} & 74\textsubscript{$\pm$2} \\
student representations \cite{lee2023debiased}   & 66\textsubscript{$\pm$2} & 91\textsubscript{$\pm$2} & 16\textsubscript{$\pm$1} & 59\textsubscript{$\pm$2} & 68\textsubscript{$\pm$1} & 67\textsubscript{$\pm$2} & 72\textsubscript{$\pm$2} \\
student outputs \cite{hinton2015distilling} & 68\textsubscript{$\pm$2} & 93\textsubscript{$\pm$2} & 30\textsubscript{$\pm$2} & 67\textsubscript{$\pm$2} & 69\textsubscript{$\pm$1} & 73\textsubscript{$\pm$2} & 78\textsubscript{$\pm$2} \\
\midrule
\textbf{TEA student (ours)} & 59\textsubscript{$\pm$2} & 89\textsubscript{$\pm$2} & 75\textsubscript{$\pm$1} & 89\textsubscript{$\pm$1} & 85\textsubscript{$\pm$1} & 93\textsubscript{$\pm$1} & 90\textsubscript{$\pm$1} \\
\bottomrule
\end{tabular*}}
\begin{tablenotes}[flushleft]\footnotesize
      \item $\dagger$ For fairness, TEA and all distillation methods use the same teacher (the segmentation–classification pipeline) and the same student architecture (ResNet50 classifier). However, the original papers for the alternative distillation techniques did not propose distilling from a segmentation-classification pipeline.
      \item $\dagger\dagger$ The teacher uses a large segmenter even at inference. Without this, its average F1-Score drops to 67\textsubscript{$\pm$1}. 
    \end{tablenotes}
\end{table}


The TEA student generalizes to unseen hospitals better than 10 state-of-the-art methods\footnote{In the X-ray dataset, we did not implement the 3 group robustness methods in Tab. \ref{tab:coco_Waterbirds_results}, because this dataset does not present separated data groups with different biases. Instead, the X-ray dataset has different source hospitals for each class: pneumonia, COVID-19 and normal. Since each source has a single class, they are not adequate groups for group robustness.} (Tab. \ref{tab:x_ray_results}), due to its higher resistance to background bias. Our training dataset has strong bias, since each class (COVID-19, pneumonia or normal) comes from a different hospital, whose X-rays have different background features \cite{bassi2024improving}. To evaluate resistance to background bias, we tested on an out-of-distribution (OOD) dataset (also from \cite{bassi2024improving}), representing hospitals never seen during training (see Supp. Note \ref{app:dataset_details} for dataset details). It is known that models influenced by bias under-perform on OOD data, which usually does not contain the same biases as the training data \cite{geirhos2020shortcut,bassi2024improving,degrave2021ai}. 
Accordingly, a standard classifier had low F1-Score in the OOD dataset (55\%, Tab. \ref{tab:x_ray_results}), and its explanation heatmaps consistently showed background attention (Fig. \ref{fig:method}). This finding demonstrates that the background bias in this X-ray training dataset is strong and detrimental to classifiers, as already shown in the paper presenting the dataset \cite{bassi2024improving}.


Like in the COCO-on-Places and Waterbirds experiments, the TEA student was resistant to background bias: its LRP heatmaps show no background attention to bias (Fig. \ref{fig:method}) and its F1-Score surpassed a standard classifier by 15\% in the OOD test dataset. 
Among all state-of-the-art methods (Tab. \ref{tab:x_ray_results}), the ISNet \cite{bassi2024improving}, the teacher (segmentation-classification pipeline) and our TEA student considerably surpassed other classifiers in generalization performances in the OOD test dataset, with 11 to 41\% higher F1-Scores (Tab. \ref{tab:x_ray_results}). The TEA student matched its teacher in the X-ray dataset, in both average F1-Score and AUC. As in COCO-on-Places and Waterbirds, the teacher here relies on a foreground segmenter both in training and in inference. The TEA student does not need a segmenter and does not add extra computational cost at inference. Thus, it is 5 $\times$ smaller and 108\% faster than the teacher. Most importantly, as in COCO-on-Places and Waterbirds, TEA surpassed state-of-the-art distillation techniques in transferring background resistance to bias from the teacher to the student in the X-ray dataset, resulting in less attention to background bias and better generalization to unseen hospitals (Tab. \ref{tab:x_ray_results}). These results confirm that training with explanations \textit{alone} results in superior resistance to background bias---both in synthetically biased photographs and real-world X-rays.

The ISNet, our recent background attention minimization method, is known to significantly surpass segmentation-classification pipelines in our X-ray dataset \cite{bassi2024improving}. Thus, it significantly surpassed the teacher (a segmentation-classification pipeline) and, consequently, the TEA student. Unlike the ISNet, the classifier inside the pipeline can pay some attention to the border of the lungs (foreground), which are apparent after lung segmentation \cite{bassi2024improving}. To make them less apparent, we added random noise to the removed X-ray background before the classification (Sec. \ref{sec:teacher}). This improved the pipeline F1-Score by 2.5\% and reduced border attention in heatmaps (Fig. \ref{fig:method}). Although the ISNet is very successful in X-ray datasets \cite{bassi2024improving}, the TEA student strongly outperformed it in COCO-on-Places and Waterbirds (Tab. \ref{tab:coco_Waterbirds_results}). The reason is that the ISNet---like all other background attention minimization methods (e.g., RRR \cite{ross2017right} and GAIN \cite{li2018tell})---has two opposing loss functions. One is the heatmap-based loss function, which avoids background attention, and the other is a standard classification loss, which indirectly prompts background attention to bias\footnote{Bias is correlated to the classes in a classification task \cite{geirhos2020shortcut,lapuschkin2019unmasking}. Thus, attention to bias may reduce the classification loss. In other words, the classification loss can promote attention to bias. Standard distillation losses can also promote attention to bias: correct teacher outputs are correlated to the classification labels, which are correlated to bias. Thus, attention to bias may help a classifier reduce standard distillation losses.}. The background bias is less extensive in X-ray datasets than in COCO-on-Places or Waterbirds\footnote{In X-rays background biases are smaller and less noticeable, and do not present colors (the image is grayscale). Example of these biases are words, hospital markers, and even contours of the patients' body \cite{bassi2024improving}.}, making the loss functions less opposing and allowing them to simultaneously converge. However, with the extensive background bias in Waterbirds and COCO-on-Places, the opposing loss functions did not effectively converge even after comprehensive hyper-parameter tuning and balancing (Supp Note \ref{app:training_augmentation}).
The TEA student has no opposing loss issue (single loss), and it converged to an accurate, bias-resistant solution in all our experiments.

Although the performances of all methods in Tab. \ref{tab:x_ray_results} may seem lower than many COVID-19 detection studies showing near-perfect F1-Scores on IID datasets, those exceptional results often stem from bias and shortcut learning---which artificially boost IID performance but hamper real-world generalization \cite{degrave2021ai,lopez2021current}. As discussed in the paper presenting our training and testing datasets \cite{bassi2024improving}, the performances seen in this test dataset match other studies using test data from unseen hospitals (OOD) \cite{degrave2021ai,bassi2024improving,bassi2022covid,lopez2021current}.

\subsection{TEA Resists Foreground Bias}
\label{sec:foreground_results}

\begin{table}[t]
\centering
\caption{\textbf{The foreground bias in the Coloured MNIST and DogsWithTies datasets had minimal influence on TEA student, which surpasses the 12 state-of-the-art methods tested.} In the table, low generalization gap indicates high resistance to foreground bias. The TEA student gap is smaller than others (p$\leq$0.05), even rivaling its teacher. Notably, the teacher is a ResNet50 trained in an unbiased version of MNIST (for the Coloured MNIST experiments) or a zero-shot CLIP \cite{radford2021learning} (for DogsWithTies), while the TEA student is a small ResNet18 trained in a highly biased dataset (Coloured MNIST/DogsWithTies). In both datasets the foreground has color and/or shape bias: each digit in Coloured MNIST is spuriously associated to a color, and each dog breed in DogsWithTies has a tie with a characteristic color and shape. For each dataset, we have three test sets. \textbf{(I)} Biased test set: all test images have biases like in training (e.g., digit 1 is blue in training and in biased test sets). \textbf{(II)} Unbiased test set: no bias, i.e., digits with random colors and dogs without ties.  \textbf{(III)} Deceiving bias test set: the correlation between foreground biases and classes is switched to deceive classifiers (e.g., the digit 1 is blue in training, 2 is blue in testing). Generalization gap indicates the difference in accuracy between the biased and deceiving test sets, measuring resistance to bias. The table shows accuracy as mean +/-std (see Supp. Note \ref{app:statistical} for details on the statistical analysis). For group robustness, 80\% of the training images had foreground bias. For all other models, including TEA, 100\% of the training images had foreground bias---a very difficult setting. The intermediate representation distillation method (student rep.) had a gap similar to TEA in DogsWithTies, but only because it did not converge and resorted to random guessing (accuracy near 50\%). Test se sizes are N=10,000 in Coloured MNIST and N=101 in DogsWithTies (justifying its wider confidence intervals).}
\label{tab:foreground}
\scriptsize
\begin{tabular}{%
  p{0.25\textwidth}                               
  *{4}{>{\centering\arraybackslash}p{0.087\textwidth}} 
  *{4}{>{\centering\arraybackslash}p{0.087\textwidth}} 
}
\toprule
 & \multicolumn{4}{c}{Coloured MNIST} &
   \multicolumn{4}{c}{DogsWithTies}\\
\cmidrule(lr){2-5}\cmidrule(lr){6-9}
\textbf{method} &
biased\ acc & unbiased\ acc & deceiv. bias\ acc & gen.\ gap $\downarrow$ &
biased\ acc & unbiased\ acc & deceiv. bias\ acc & gen.\ gap $\downarrow$  \\
\midrule
AG-Sononet \cite{schlemper2018attention}            & 100\textsubscript{$\pm0$} & 16\textsubscript{$\pm0$} & 0\textsubscript{$\pm0$} & 100\textsubscript{$\pm0$} &
                         99\textsubscript{$\pm2$} & 50\textsubscript{$\pm1$} & 3\textsubscript{$\pm2$} & 96\textsubscript{$\pm3$} \\
ViT-B/16 \cite{dosovitskiy2020image}              & 100\textsubscript{$\pm0$} & 21\textsubscript{$\pm0$} & 1\textsubscript{$\pm0$} & 99\textsubscript{$\pm0$} &
                        88\textsubscript{$\pm3$} & 58\textsubscript{$\pm4$} & 52\textsubscript{$\pm5$} & 36\textsubscript{$\pm6$} \\
standard classifier \cite{he2016deep}     & 100\textsubscript{$\pm0$} & 18\textsubscript{$\pm0$} & 0\textsubscript{$\pm0$} & 100\textsubscript{$\pm0$} &
                        98\textsubscript{$\pm2$} & 69\textsubscript{$\pm4$} & 41\textsubscript{$\pm5$} & 57\textsubscript{$\pm5$} \\
\midrule
\multicolumn{9}{l}{\textit{group robustness}}\\
IRM \cite{arjovsky2019invariant}                  & 100\textsubscript{$\pm0$} & 60\textsubscript{$\pm1$} & 53\textsubscript{$\pm1$} & 47\textsubscript{$\pm1$} &
                        85\textsubscript{$\pm4$} & 67\textsubscript{$\pm4$} & 27\textsubscript{$\pm4$} & 57\textsubscript{$\pm5$} \\
GroupDRO \cite{sagawa2019distributionally}             & 100\textsubscript{$\pm0$} & 52\textsubscript{$\pm2$} & 40\textsubscript{$\pm2$} & 59\textsubscript{$\pm2$} &
                        71\textsubscript{$\pm4$} & 70\textsubscript{$\pm4$} & 63\textsubscript{$\pm5$} & 8\textsubscript{$\pm6$} \\
PGI \cite{ahmed2020systematic}                  & 100\textsubscript{$\pm0$} & 64\textsubscript{$\pm2$} & 58\textsubscript{$\pm2$} & 42\textsubscript{$\pm2$} &
                        88\textsubscript{$\pm3$} & 71\textsubscript{$\pm4$} & 56\textsubscript{$\pm5$} & 32\textsubscript{$\pm6$} \\
\midrule
\multicolumn{9}{l}{\textit{distillation}}\\
TEA teacher$^\dagger$   &  99\textsubscript{$\pm0$} & 99\textsubscript{$\pm0$} & 99\textsubscript{$\pm0$} & 0\textsubscript{$\pm0$}  &
                        96\textsubscript{$\pm2$} & 96\textsubscript{$\pm2$} & 96\textsubscript{$\pm2$} & 0\textsubscript{$\pm3$} \\
st. G.-CAM \& out. \cite{parchami2024good}      & 100\textsubscript{$\pm0$} & 91\textsubscript{$\pm0$} & 85\textsubscript{$\pm0$} & 15\textsubscript{$\pm0$}   &
                        75\textsubscript{$\pm4$} & 72\textsubscript{$\pm4$} & 70\textsubscript{$\pm4$} & 5\textsubscript{$\pm5$} \\
student rep. \cite{lee2023debiased}        & 100\textsubscript{$\pm0$} & 92\textsubscript{$\pm0$} & 87\textsubscript{$\pm0$} & 13\textsubscript{$\pm0$} &
                        53\textsubscript{$\pm4$} & 53\textsubscript{$\pm4$} & 55\textsubscript{$\pm3$} & 2\textsubscript{$\pm5$} \\
student outputs \cite{hinton2015distilling}       & 100\textsubscript{$\pm0$} & 93\textsubscript{$\pm0$} & 87\textsubscript{$\pm0$} & 13\textsubscript{$\pm0$} &
                        87\textsubscript{$\pm3$} & 81\textsubscript{$\pm3$} & 76\textsubscript{$\pm4$} & 11\textsubscript{$\pm5$} \\
GALS \cite{GALS}\textsuperscript{$\dagger\dagger$}                   & N/A   & N/A   & N/A   & N/A   &
                        76\textsubscript{$\pm4$} & 74\textsubscript{$\pm4$} & 72\textsubscript{$\pm4$} & 4\textsubscript{$\pm6$} \\
\midrule
\textbf{TEA student (ours)}     & 99\textsubscript{$\pm0$} & 98\textsubscript{$\pm0$} & 98\textsubscript{$\pm0$} & 1\textsubscript{$\pm0$} &
                        79\textsubscript{$\pm4$} & 82\textsubscript{$\pm4$} & 78\textsubscript{$\pm4$} & 2\textsubscript{$\pm5$} \\
\bottomrule
\end{tabular}
\begin{tablenotes}[flushleft]\footnotesize
      \item $\dagger$ Teacher is trained on datasets w/o the foreground bias, unlike the students. The teacher is a ResNet50 trained on a random Coloured MNIST or a zero-shot CLIP (ResNet50 backbone). Students are ResNet18.
      \item $\dagger\dagger$ GALS is based on learning a student from CLIP. Thus, it was only applied to the DogsWithTies dataset, where we used CLIP as the teacher.
\end{tablenotes}
\end{table}


The TEA student surpassed all 11 implemented state-of-the-art methods in resistance to foreground bias (Tab. \ref{tab:foreground}). In this study, foreground bias refers to colors or objects inside the image's region of interest. In the Coloured MNIST dataset, we classify 10 hand-written digits, each with a different color (bias). In DogsWithTies (our biased version of StandordDogs \cite{dataset2011novel}, see Supp. Note \ref{app:dataset_details}), we classify the Pekingese and Tibetan Mastiff dog breeds, but Tibetan Mastiffs have red neckties and Pekingese have purple bow ties (bias). As in COCO-on-Places and Waterbirds, the biases are synthetically added and controllable. Thus, we created biased, unbiased and deceiving bias test sets, and calculated a generalization gap (biased test set accuracy minus deceiving bias accuracy) to quantitatively measure the influence of background bias over the classifiers. The TEA student had the smallest generalization gaps, 1\% in Coloured MNIST and 2\% in DogsWithTies. Naturally, we did not compare against background attention minimization methods (ISNet, GAIN or RRR) or the multi-task U-Net, as they cannot be used to mitigate foreground bias, unlike TEA.

To mitigate foreground bias, the explanation heatmaps used to train the TEA student come from teachers that are resistant to the biases in Colored MNIST and DogsWithTies. These teachers are a ResNet-50 trained on a randomly colored MNIST, and a zero-shot CLIP (ResNet-50 variant). The TEA student is a compact classifier, the ResNet-18.
Our demonstrative experiments simulate a situation where a researcher has both a biased dataset and a large teacher trained elsewhere, using an inaccessible dataset without the bias present in the researcher's dataset. TEA compresses such a teacher into a smaller student architecture while preserving resistance to bias, whereas standard distillation techniques create biased students (Tab. \ref{tab:foreground}). Directly training the small architecture on the teacher’s data would be simpler, but many modern AI models are released without their training datasets---normally due to commercial interests or patient privacy concerns and regulations. Examples are CLIP \cite{radford2021learning} (foundational vision-language model, OpenAI), BiomedCLIP \cite{zhang2023biomedclip} (foundational biomedical vision-language model, Microsoft), Llama 3.0 \cite{dubey2024llama} (foundational large language model, Meta), and RETFound \cite{zhou2023foundation} (foundational model for retinal images, University College London). Naturally, public AI models are not immune to all types of bias \cite{baherwani2024racial}. Before using a public classifier as teacher, we suggest using OOD test datasets and explanation heatmaps to investigate the classifier resistance to biases relevant for the researcher. E.g., CLIP resists the ties bias in DogsWithTies, but is subject to gender bias \cite{baherwani2024racial}. If a suitable teacher is found, TEA is a practical way to downsize it using any available---and potentially biased---data (Tab. \ref{tab:foreground}).

Intermediate representation distillation \cite{lee2023debiased}, which was very promising in Tab. \ref{tab:coco_Waterbirds_results}, completely failed on DogsWithTies (accuracy near 50\%, Tab. \ref{tab:foreground}). This failure indicates that distilling from the CLIP teacher’s intermediate representations with a small dataset (N=180) is difficult. CLIP is a multi-purpose model trained on a massive, diverse dataset \cite{radford2021learning}, so its intermediate representations capture a wide range of image information. A student trained on only a few hundred dog images may therefore struggle to reproduce these complex representations. In contrast, TEA only requires the student to learn the CLIP teacher’s explanation heatmaps (LRP). These heatmaps capture how each image feature influences the CLIP teacher’s dog breed classification output. Thus, they focus on the image information that is relevant to the dog breed classification task only. By distilling explanation heatmaps, TEA provides a much easier and effective target for the student. Therefore, the TEA student converged to high accuracy, but the student in representation distillation did not converge.
 

\subsection{It is Better to Train with Explanations Alone}
\label{sec:ablations_results}

\begin{table}[!h]
\centering
\caption{\textbf{Ablation studies: training with explanations \textit{alone} (TE\textit{A}) surpasses training with explanations plus other forms of distillation; and TEA is more bias resistant with LRP explanation heatmaps.} In the table, low generalization gap and high accuracy on the unbiased and deceiving bias test sets indicate high resistance to bias. We perform multiple ablation studies on the COCO-on-Places datasets. \textit{Training with explanations, alone or not alone?} adding other distillation losses to TEA (distilling outputs / intermediate representations) reduces its resistance to bias. Thus, classifiers leveraged bias and shortcut learning to minimize these additional distillation losses. \textit{Explanation heatmap ablations.} Substituting LRP-$\varepsilon$ explanation heatmaps by GradCAM, Gradient*Input, input gradients or attention also reduced the TEA resistance to bias.  \textit{$\varepsilon$ hyper-parameter ablations.} The best resistance to bias was achieved with LRP-$\varepsilon$ and $\varepsilon$ randomly chosen between 0.01 and 0.001 for each training image (standard TEA). Still, the common choice of $\varepsilon=0.01$ \cite{bassi2024improving} also yields strong results, and we see no need of tuning $\varepsilon$. \textit{Loss ablations.} Substituting our custom multi-resolution TEA Loss by its single-resolution version (Sec. \ref{sec:loss}) or simple L1 / L2 losses also reduced resistance to bias.
The table shows accuracy as mean +/-std (see Supp. Note \ref{app:statistical} for details on the statistical analysis). 
\textbf{(I)} Biased test set: all test images have biases like in training. \textbf{(II)} Unbiased test set: no bias.  \textbf{(III)} Deceiving bias test set: the correlation between foreground biases and classes is switched to deceive classifiers. Classifiers are ResNet50.}
\label{tab:ablations}

\scriptsize
\begin{tabular}{%
  p{0.5\linewidth}                               
  *{4}{>{\centering\arraybackslash}p{0.115\linewidth}} 
}
\toprule
 & \multicolumn{4}{c}{\textbf{COCO-on-Places}} \\
\cmidrule(lr){2-5}
\textbf{method} &
biased\ acc & unbiased\ acc & deceiving bias\ acc & generaliz.\ gap $\downarrow$ \\
\midrule
standard classifier  \cite{he2016deep} & 73\textsubscript{$\pm$1} & 21\textsubscript{$\pm$1} &  10\textsubscript{$\pm$1} & 64\textsubscript{$\pm$1} \\
TEA teacher & 72\textsubscript{$\pm1$} & 72\textsubscript{$\pm1$} & 72\textsubscript{$\pm1$} & 1\textsubscript{$\pm2$} \\
\midrule
\multicolumn{5}{l}{\textit{ablations: explanation heatmaps}}\\
TEA Gradient*Input   & 25\textsubscript{$\pm1$} & 17\textsubscript{$\pm1$} & 15\textsubscript{$\pm1$} & 10\textsubscript{$\pm2$} \\
TEA Input Gradient   & 28\textsubscript{$\pm1$} & 18\textsubscript{$\pm1$} & 16\textsubscript{$\pm1$} & 12\textsubscript{$\pm2$} \\
TEA Grad-CAM         & 67\textsubscript{$\pm1$} & 30\textsubscript{$\pm1$} & 24\textsubscript{$\pm1$} & 43\textsubscript{$\pm2$} \\
TEA Attention        & 83\textsubscript{$\pm1$} & 62\textsubscript{$\pm2$} & 57\textsubscript{$\pm2$} & 26\textsubscript{$\pm2$} \\
\midrule
\multicolumn{5}{l}{\textit{ablations: $\varepsilon$ hyper-parameter in LRP-$\varepsilon$}}\\
TEA LRP-$\varepsilon=10^{-1}$  & 79\textsubscript{$\pm1$} & 68\textsubscript{$\pm2$} & 65\textsubscript{$\pm2$} & 14\textsubscript{$\pm2$}  \\
TEA LRP-$\varepsilon=10^{-2}$  &  76\textsubscript{$\pm1$} & 71\textsubscript{$\pm1$} & 67\textsubscript{$\pm2$} & 8\textsubscript{$\pm2$}  \\
TEA LRP-$\varepsilon=10^{-3}$  &  72\textsubscript{$\pm1$} & 68\textsubscript{$\pm2$} & 67\textsubscript{$\pm2$} & 5\textsubscript{$\pm2$}  \\
TEA LRP-$\varepsilon=10^{-4}$  & 50\textsubscript{$\pm2$} & 45\textsubscript{$\pm2$} & 10\textsubscript{$\pm1$} & 40\textsubscript{$\pm2$}  \\
TEA LRP-$\varepsilon=10^{-5}$  & 12\textsubscript{$\pm1$} & 11\textsubscript{$\pm1$} & 14\textsubscript{$\pm1$} & 2\textsubscript{$\pm1$}  \\
\midrule
\multicolumn{5}{l}{\textit{ablations: heatmap distillation losses}}\\
TEA LRP-$\varepsilon$ L2 & 42\textsubscript{$\pm2$} & 38\textsubscript{$\pm2$} & 32\textsubscript{$\pm1$} & 10\textsubscript{$\pm2$} \\
TEA LRP-$\varepsilon$ L1 & 75\textsubscript{$\pm1$} & 69\textsubscript{$\pm2$} & 67\textsubscript{$\pm2$} & 8\textsubscript{$\pm2$}  \\
TEA LRP-$\varepsilon$ single-resolution & 73\textsubscript{$\pm1$} & 67\textsubscript{$\pm2$} & 66\textsubscript{$\pm2$} & 7\textsubscript{$\pm2$}  \\
\midrule
\multicolumn{5}{l}{\textit{ablations: Training with Explanations \textbf{Alone} or not alone}}\\
TEA LRP-$\varepsilon$ + output distillation &  79\textsubscript{$\pm1$} & 66\textsubscript{$\pm2$} & 64\textsubscript{$\pm2$} & 15\textsubscript{$\pm2$} \\
TEA LRP-$\varepsilon$ + representation distillation & 79\textsubscript{$\pm1$} & 68\textsubscript{$\pm1$} & 65\textsubscript{$\pm2$} & 14\textsubscript{$\pm2$}  \\ 
\midrule
\textbf{TEA student (ours)}\textsuperscript{$\dagger$}  & 74\textsubscript{$\pm$1} & 71\textsubscript{$\pm$2} & 69\textsubscript{$\pm$2} &  4\textsubscript{$\pm$2} \\ 
\bottomrule
\end{tabular}
\begin{tablenotes}
    \item $\dagger$ Standard TEA: LRP-$\varepsilon$, with $\varepsilon$ randomly chosen between $10^{-3}$ and $10^{-2}$.
\end{tablenotes}
\end{table}





Ablation studies in Tab. \ref{tab:ablations} (bottom) shows that training with explanations \textit{alone} leads to superior resistance to background bias. The inclusion of an output distillation loss (or intermediate representation distillation) in TEA significantly reduced its resistance to background bias. This finding is also supported by all our previous experiments (Tab. \ref{tab:coco_Waterbirds_results}, \ref{tab:x_ray_results} and \ref{tab:foreground}), which show that the TEA student is more resistant to bias than a classifier trained with the distillation of explanations (Grad-CAM) plus outputs \cite{parchami2024good}. Unlike standard students, the TEA student high accuracy is a sole consequence of learning the reasons behind the teacher outputs, thus \textit{naturally} arriving at the similar outputs.

Table \ref{tab:ablations} also shows ablation studies on the explanation heatmaps used to train the TEA student. We experiment with 5 types of explanation heatmap: LRP \cite{bach2015onpixel}, attention (feature-based spatial attention \cite{attentionTransfer}), Grad-CAM \cite{selvaraju2017grad}, input gradients \cite{simonyan2014deep} and Gradient*Input \cite{avanti2016not}.\footnote{There are many explanation techniques. We experimented with these five because they are popular, differentiable, fast, and used in background attention minimization studies (thus, they can be used inside loss functions) \cite{bassi2024improving,ross2017right,li2018tell}.}
LRP led to the highest accuracy and resistance to bias for the TEA student. These results align with previous findings in background attention minimization, where LRP surpassed other explanation heatmap types \cite{bassi2024improving}. Sec. \ref{sec:explanations} shows that Grad-CAM and attention distillation can be seen as special cases of distilling intermediate representations. These three methodologies have fundamental shortcomings, which reduce their ability to mitigate attention to bias. Section \ref{sec:explanations} also provides guidance for choosing or designing explanation techniques for TEA, and theoretically justifies the reasons \textit{why} LRP was superior. Key reasons are the LRP superior resilience to noise, especially for deep classifiers and high-resolution images, its capacity of tracing back how the influence of bias flows from the input image until the classifier outputs, and the stronger theoretical principles and guarantees LRP provides to TEA. 

Supp. Note \ref{app:epsilon_ablations} interprets ablation studies on the $\varepsilon$ hyper-parameter of LRP (Tab. \ref{tab:ablations}), and explains why we suggest there is no need to fine-tuning this hyper-parameter. Supp. Note \ref{sec:loss} analyzes the ablation studies on our custom TEA Loss (Tab. \ref{tab:ablations}), supporting the loss design. Importantly, in TEA we use LRP-Flex \cite{bassi2024faster}, an LRP implementation that can be readily applied to any ReLU-based classifier architecture in PyTorch. Thus, TEA is easy to use and to implement to new classifier architectures.

\section{Discussion}




TEA trains classifiers to focus on important image features, not bias. Its resistance to background and foreground bias consistently surpassed all 14 state-of-the-art methods in 5 datasets. In a single dataset, one method (the ISNet \cite{bassi2024improving}) surpassed TEA (Tab. \ref{tab:x_ray_results}). This is because, in this dataset, the ISNet surpassed the teacher model from which TEA learned from. However, in two other datasets (Tab. \ref{tab:coco_Waterbirds_results}), TEA strongly surpassed the ISNet, since TEA solves the problem of opposing losses present in all background attention minimization methods like the ISNet, RRR, and GAIN. Moreover, TEA can also address biases in the foreground, which is not possible for the ISNet or other background attention minimization methods. Overall, the experiments in five datasets show the versatility of TEA: it addresses foreground (Tab. \ref{tab:foreground}) and background biases (Tabs. \ref{tab:coco_Waterbirds_results}, \ref{tab:x_ray_results}); resists both extensive synthetic bias (Tabs. \ref{tab:coco_Waterbirds_results}, \ref{tab:foreground}) and bias naturally occurring in X-rays (Tab. \ref{tab:x_ray_results}); it works for both low (Tab. \ref{tab:foreground}) and high resolution (Tab. \ref{tab:x_ray_results}) images; small (Tab. \ref{tab:foreground}) and large datasets (Tab. \ref{tab:x_ray_results}); and for photographs (Tabs. \ref{tab:coco_Waterbirds_results}, \ref{tab:foreground}) and medical images (X-rays, Tab. \ref{tab:x_ray_results}). Mitigating background bias is critical---it is a risk to medical AI and was observed in hundreds of published X-ray classifiers \cite{degrave2021ai}. Therefore, background bias is the main focus of our study. However, we also considered color and object biases in the foreground. In future work, we plan to assess the TEA resistance to other biases, such as gender and social biases \cite{buolamwini2018gendershades}, lighting bias \cite{hendrycks2019robustness}, and frequency bias \cite{pmlr-v97-rahaman19a}. We also plan to explore TEA beyond computer vision.


The TEA superior resistance to bias was quantitatively demonstrated: we used test datasets with bias, without bias and with deceiving bias to calculate generalization gaps. They measure the influence of bias on the classifier outputs, and the generalization the gap for the TEA student was significantly smaller than that of state-of-the-art methods in multiple datasets (Tabs. \ref{tab:coco_Waterbirds_results} and \ref{tab:foreground}). We also showed that, by increasing the resistance to background bias in X-rays, TEA yielded better generalization to unseen hospitals in classifying pneumonia and COVID-19 (Tab. \ref{tab:x_ray_results}).
We compared TEA against a wide variety of methods, including distillation, group robustness and background attention minimization. The TEA student accuracy and resistance to background was even close to the accuracy and bias-resistance of its teacher, usually a segmentation-classification pipeline that is 108\% slower and 5 times larger than the TEA student at inference. Thus, TEA can compress large teachers into small students, while retaining resistance to bias better than state-of-the-art distillation techniques.

There is active research on improving explanation heatmaps \cite{achtibat2023attribution}, and active debate on whether we should use explanation heatmaps to explain deep learning classifiers, or move towards intrinsically explainable classifiers \cite{rudin2019stop}. This study does not fit into this debate. Instead, this study views explanation heatmaps as a \textit{powerful training constraint}: we quantitatively demonstrate that explanation heatmaps from deep classifiers can be used to constrain what these large AI models learn, avoiding shortcut learning.

Importantly, we are the first to demonstrate that it is possible to train a classifier with explanation heatmaps alone (TEA). TEA does not apply any loss function to the classifier outputs or intermediate representations. Yet, the TEA student achieves high classification accuracy by learning, through heatmaps, how image features influenced its teacher outputs. Therefore, the TEA student inherits both the teacher's accuracy and its resistance to background bias. Notably, employing alternative distillation losses together with the explanation-based TEA loss made the TEA student less resistant to bias (Tab. \ref{tab:ablations}). That is because the long-standing teacher-student paradigm \cite{vapnik2009learning,hinton2015distilling} requires a student to match the output or intermediate representations of a teacher, but the student can leverage biases and shortcut learning to achieve this goal (Tab. \ref{tab:ablations}). Conversely, \textbf{T}raining with \textbf{E}xplanations \textbf{A}lone requires the student to learn the reasons behind the teacher outputs. Thus, the student naturally arrives at outputs similar to the teacher's, instead of learning to parrot the output of the teacher.




\section{Methods}




The TEA training paradigm mitigates shortcut learning by training a classifier to focus on relevant image features---rather than background (or foreground) bias. TEA is diverse from the standard training paradigm: because bias is, by definition \cite{lapuschkin2019unmasking,geirhos2020shortcut}, correlated to standard classification labels, TEA abandons any classification (or distillation) loss function applied to the output of the classifier or to its intermediate representations. Instead, TEA trains a classifier by only optimizing the reasons behind its outputs. These reasons are represented by explanation heatmaps (e.g., LRP) that reveal how every region in the input image contributed to the classifier output (Supp. Note \ref{app:why_LRP}). TEA optimizes the classifier explanation heatmaps to match target heatmaps, using a novel loss function. These target heatmaps show no attention to background bias, being created from a teacher model that is resistant to these biases. Crucially, TEA transfers the performance and bias-resistance of large teacher models (e.g, pipelines that segment the image foreground before classifying it) to much smaller classifiers (TEA student). By learning the reasons behind the teacher outputs instead of learning to copy these outputs, the TEA student better inherits the teacher's resistance to bias. Sec. \ref{sec:teacher} discusses how to create the teacher \& target heatmaps for TEA, and Sec. \ref{sec:loss} discusses the TEA Loss Function. Section Sec. \ref{sec:explanations} theoretically justifies TEA's strong performance, and explains why LRP surpassed other 4 types of explanation heatmaps in TEA---delineating key explanation heatmaps qualities that benefit TEA. Fig. \ref{fig:method} summarizes TEA.


\subsection{Teacher \& Target Explanation Heatmaps}
\label{sec:teacher}


To resist background bias, the teacher is a segmentation-classification pipeline, trained (see training details in Supp. Note \ref{app:training_augmentation}) and frozen before training the TEA student. In this pipeline, a segmenter (e.g., U-Net) finds and removes the image background (and biases within it), and a classifier receives the image without background. Usually, the pipeline substitutes the image background by a black background \cite{bassi2022covid}. However, this can make the background borders very apparent, and the classifier inside the pipeline can pay excessive attention to them, slightly overfitting to the borders' shape and position \cite{bassi2024improving}. To mitigate this problem, we substitute the black background with uniform random noise, making borders less apparent and improving the generalization of the pipeline (Sec. \ref{sec:background_results_x_ray}). Importantly, the classifier inside the segmentation-classification pipeline loses accuracy if we use it without the segmenter (Tab. \ref{tab:coco_Waterbirds_results}), making the pipeline much larger and slower than a standalone classifier at inference. TEA transfers the bias-resistance of this large segmentation-classification pipeline to the much lighter TEA student, with no segmenter. 

From the teacher we create the target heatmaps, $\mathbf{H_{t}}$, used to train the TEA student. When training the TEA student, we send each training image to both the teacher and the TEA student. They classify the image, and we create the respective explanation heatmaps, $\mathbf{H_{t}}$ and $\mathbf{H_{s}}$. To create $\mathbf{H_{t}}$, the explanation technique (e.g., LRP) is only applied to the classifier inside the segmentation-classification pipeline, not to the segmenter. Therefore, $\mathbf{H_{s}}$ and $\mathbf{H_{t}}$ are a pair of explanation heatmaps for two different---but spatially aligned---versions of the same input image: one with background (TEA student) and one without background (teacher).

\subsubsection{Extension to Foreground Bias}



The TEA main focus is resisting background bias, but we also extended it to resist color and object biases in the image foreground. In this case, the segmentation-classification pipeline cannot be used as teacher, because background segmentation cannot remove foreground biases. However, when we have a training dataset (A) with a foreground bias, a classifier trained in a different dataset (B) may better resist the foreground bias in dataset A. In this case, an ideal solution is to train on dataset B. But this is not always possible, as there is an increasing number of studies that release large trained classifiers (e.g., foundational models), but not their training datasets \cite{radford2021learning,zhang2023biomedclip,dubey2024llama,zhou2023foundation}. In our demonstrative experiments (Sec. \ref{sec:foreground_results}), we have a training dataset (A) about dog breed classification, where dogs have ties as foreground bias. Our teacher is CLIP \cite{radford2021learning}, a large foundational model, trained on a dataset (B) where, naturally, most dogs do not have ties. It is not possible to train on CLIP's dataset (B) because it is too massive, and it is private. However, TEA can train a small classifier on our biased training dataset (B), inheriting CLIP's resistance to the foreground bias (ties). I.e., TEA can transfer the foreground resistance to bias of large public classifiers to a small---easy to deploy---TEA student. Before using a public classifier as teacher, we suggest using explanation heatmaps or OOD evaluation to assess its resistance to the specific foreground biases present in the TEA training dataset.

\subsection{TEA Loss Function}
\label{sec:loss}

The TEA Loss Function, $\mathcal{L}_{\mathrm{tea}}(\mathbf{H_{s}},\mathbf{H_{t}})$, is the only loss function minimized in the TEA training paradigm. It calculates a dissimilarity metric between the student explanation heatmap, $\mathbf{H_{s}}$, and the target heatmap, $\mathbf{H_{t}}$. Both are 3D tensors, with shape C, H, W, where C is the number of channels (3 in RGB images), and H and W are the height and width of the input image, respectively. We designed this metric to address key challenges in training with explanation heatmaps: noise, and large norm differences of $\mathbf{H_{s}}$ and $\mathbf{H_{t}}$. Here, we present the TEA Loss and its design principles, which are empirically supported in Supp. Note \ref{sec:loss_ablations}. To compute TEA Loss, we first calculate $d(\mathbf{H_{s}},\mathbf{H_{t}})$, the L1 distance between normalized teacher and the student heatmaps:

\begin{gather}
    \label{eq:dist}
    d(\mathbf{H_{s}},\mathbf{H_{t}})= \frac{||\mathbf{H_{s}}-\mathbf{H_{t}}||_{1}}{\sqrt{||\mathbf{H_{s}}||_{1}  ||\mathbf{H_{t}}||_{1}}}
\end{gather}

L1-based losses are more resistant to outliers than L2-based losses, improving the convergence and performance of the TEA student. The norm of explanation heatmaps depends on training and architecture. Thus, the TEA student and the target explanation heatmaps may have norms at different scales in the beginning of training. To avoid norm differences to dominate the TEA Loss, or cause training instability, we normalize both $\mathbf{H_{s}}$ and $\mathbf{H_{t}}$ before calculating the L1 distance. We divide them by the geometric mean of their L1 norms (Eq. \ref{eq:dist}). Conversely, individual normalization (dividing each each heatmap by its norm) is not ideal here. After individual normalization, the L1 loss will not enforce the norms of the two explanation heatmaps to match\footnote{The L1 loss with individual normalization is defined as $||(\mathbf{H_{s}}/||\mathbf{H_{s}}||_{1})-(\mathbf{H_{t}}/||\mathbf{H_{t}}||_{1})||_{1}$. It is zero when $\mathbf{H_{t}}=\alpha\mathbf{H_{s}} \forall \alpha \in \Re^{+}$.}, but explanation heatmap norms carry meaningful information, such as the classifier confidence in its decisions. The TEA loss normalization avoids training instability, but still enforces the TEA student explanation heatmaps to exactly match. Finally, to improve the TEA resistance to heatmap noise, we formulate the TEA Loss as a ``multi-resolution loss'':

\begin{gather}
    \label{eq:pyramid}
    \mathcal{L}_{\mathrm{tea}}(\mathbf{H_{s}},\mathbf{H_{t}})=\frac{1}{M}\sum_{m=0}^{M-1} d(\mathrm{AvgPool}(\mathbf{H_{s}}; K=2^{m}),\mathrm{AvgPool}(\mathbf{H_{t}};K=2^{m}))
\end{gather}

Explanation heatmaps can be noisy, especially for deep classifiers \cite{ancona2017towards}. Average pooling (2D), AvgPool$(\cdot)$, can reduce high-frequency noise by averaging together the explanation  heatmap values for nearby pixels. On the other hand, pooling reduces the resolution of heatmaps, losing fine details. The TEA Loss uses multiple average pooling operations, with kernel sizes varying from $K=1$ (original high-resolution explanation heatmap) to $K=M-1$\footnote{We empirically set the M hyper-parameter to make the size of the smallest low-resolution explanation heatmap 8 by 8}, to create low-resolution versions of the teacher and student heatmaps (Eq. \ref{eq:pyramid}). For each resolution, we calculate the dissimilarity $d(\cdot,\cdot)$ between corresponding TEA student and target heatmaps, and we average the results (Eq. \ref{eq:pyramid}). Therefore, the TEA Loss leverages high-resolution explanation heatmaps to achieve fine control over the classifier, and low-resolution (but less noisy) explanation heatmaps to improve training convergence and stability.


\subsection{TEA Theoretical Fundamentals \& Explanation Techniques}
\label{sec:explanations}


A main reason for the high accuracy and resistance to bias of TEA is the sole use of explanation heatmaps to train a classifier. Indeed, Tab. \ref{tab:ablations} (bottom) shows that resistance to bias drops if we include standard distillation losses in TEA (distillation of outputs or intermediate representations).
By training with explanation heatmaps alone, the TEA student learns the reasons behind the teacher outputs, thus ignoring bias is if the teacher also ignores it. Multiple explanation techniques (the techniques that create explanation heatmaps) exist, and improving them is an active research field \cite{achtibat2023attribution,qiu2023fgcam,selvaraju2017grad,byun2023vitreciprocam,lapuschkin2019unmasking}. We present TEA as a general training paradigm, which can use different explanation techniques. Here, we experimented with five: Layer-wise Relevance Propagation (LRP) \cite{bach2015onpixel}, Grad-CAM \cite{selvaraju2017grad}, Input Gradients \cite{simonyan2014deep}, Gradient*Input \cite{avanti2016not}, and attention \cite{attentionTransfer}. TEA requires explanation techniques that are fast and create differentiable explanation heatmaps. The five aforementioned explanation techniques meet these criteria\footnote{Differentiability is needed because gradients must back-propagate through heatmaps in training; fast techniques are important to keep training time manageable (e.g., with LRP, training takes two days with one GPU for our X-ray experiments, see Supp. Note \ref{app:speed}). LRP can create multiple heatmaps for one image, each explaining one classification class \cite{bach2015onpixel}. To reduce TEA training time, we create only two heatmaps for each training image (the student and teacher explanation heatmaps, $\mathbf{H_{s}}$ and $\mathbf{H_{t}}$), both explaining the same class, which is randomly chosen. The winning class (the class the teacher predicted) gets 50\% of probability of being chosen. The heatmap for the winning class is important, because it typically contains relevant information on the reasons why the teacher predicted that class \cite{bach2015onpixel}. However, the TEA student must also learn the reasons why it should \textit{not} predict the remaining (losing) classes. Information on these reasons is mostly found in heatmaps explaining these losing classes \cite{bach2015onpixel}. Therefore, we randomly pick one of the losing classes with equal probability, (50/(C – 1))\%, where C is the number of classes.}. Tab. \ref{tab:ablations} shows that the choice of the explanation technique has a major impact on the TEA student performance---LRP outperformed all other explanation techniques. Therefore, in all other experiments (Tab. \ref{tab:coco_Waterbirds_results}-\ref{tab:foreground}) we used LRP in TEA. We used the LRP-Flex implementation of LRP \cite{bassi2024faster}.

Here, we analyze three key LRP qualities that makes it a strong candidate for supporting TEA, with respect to the other explanation techniques: \textit{full back-propagation, guarantee of correct outputs, and resistance to noise}. We expect the following analysis to help readers choose---or create---explanation techniques that may improve the performance of TEA even further.

\textbf{Principled, full backpropagation: tracing back the influence of bias.} To create an explanation heatmap, LRP back-propagates a signal, called \textit{relevance}, through the entire classifier. This procedure gathers information from all classifier layers, transferring to the explanation heatmap both the high-level of abstraction of deep layers, and the detailed spatial resolution of early layers. TEA uses LRP-$\varepsilon$---a relevance back-propagation rule theoretically principled in efficient approximations of Taylor expansions \cite{LRPBook}. A Taylor expansion can estimate how much each input of a neuron in the classifier influenced the neuron's output \cite{LRPBook}. LRP-$\varepsilon$ back-propagates the relevance from the neuron output to its inputs proportionally to this influence \cite{LRPBook}. Relevance is back-propagated layer-by-layer, through the entire classifier. For background bias to influence the outputs of a classifier, its influence must flow through all classifier layers. In this case, LRP will produce a corresponding flow of relevance, from the classifier output until the bias location in the LRP heatmap. I.e., LRP traces back the influence of background bias. In TEA, the LRP heatmaps of the TEA student are optimized to match target heatmaps bearing minimal relevance in background or foreground bias. Consequently,  TEA minimizes the influence flowing from background/foreground bias to the student output. Grad-CAM and attention explanation heatmaps are not based on a full backpropagation process like LRP. Thus, minimization of attention to bias in these heatmaps may not minimize the influence of bias on the classifier \cite{bassi2024improving}. Accordingly, TEA with LRP surpassed TEA with Grad-CAM or attention explanation heatmaps (Tab. \ref{tab:ablations}). Supp. Note \ref{app:why_LRP} details why distilling Grad-CAM and attention heatmaps is a special case of distilling intermediate representations. Moreover, it explains why these three types of distillation may become unreliable and unable to suppress attention to bias (spurious mapping)---unlike LRP distillation. Input gradients and Gradient*Input are theoretically principled explanation heatmaps based on full back-propagation, as LRP. However, they are noisier for deep classifiers and high-resolution images (see ``resistance to noise'' below).

\textbf{Guarantee of correct outputs: by learning the reasons behind the teacher outputs, the TEA student outputs become similar to the teacher's.} 
It is reasonable to expect that, if two classifiers give the same explanation heatmaps (for the same image), then their corresponding outputs should match. This expectation is approximately guaranteed in LRP heatmaps (Supp. Note \ref{app:why_LRP}) and Input*Gradient, but not in the other types of explanation heatmaps we tested. Therefore, LRP heatmaps provides TEA a ``guarantee of correct outputs'': if the TEA student LRP heatmap matches the teacher LRP heatmap, the student output will also match the teacher's output (Supp. Note \ref{app:why_LRP}). 
Note that matching LRP heatmaps approximately guarantees matching outputs, but matching outputs \textit{does not} guarantee matching LRP heatmaps---the same output can be explained by many different LRP heatmaps (some may show attention to bias, some may not). Therefore, optimizing student outputs to match teacher outputs (regular output distillation) \textit{by no means} guarantees that LRP explanation heatmaps for the teacher and student will match. Accordingly, our experiments show that output distillation can create students that pay attention to bias, even when the teacher---and its LRP heatmaps---show no attention to it. Conversely, TEA is a way more constrained form of optimization than output distillation, making the student pay attention to the same image features that the teacher pays attention to.

\textbf{Resistance to noise: improving the training convergence of TEA for deep classifiers and high-resolution images.} Noise in explanation heatmaps can make the TEA loss function noisy, hindering its convergence and consequently limiting the accuracy of the TEA student. LRP-$\varepsilon$ has a hyper-parameter, $\varepsilon$, used to reduce noise in explanation heatmaps. In summary, larger $\varepsilon$ reduces noise by reducing the effect of neurons with weak activations on the explanation heatmaps \cite{bassi2024improving,bassi2024faster}. However, if $\varepsilon$ is too high, relevant information may be removed from the heatmaps---such as a weak influence of background bias on the classifier. Our experiments show ideal $\varepsilon$ around 0.01 (Tab. \ref{tab:ablations}). This is the same value suggested in our previous attention background minimization study \cite{bassi2024improving}, the ISNet---demonstrating there is no need to carefully tune the $\varepsilon$ hyper-parameter. When $\varepsilon=0$, LRP-$\varepsilon$ is equivalent to Gradient*Input, which is noisy for deep classifiers analyzing high-resolution images \cite{ancona2017towards,bassi2024improving}. Accordingly, TEA with LRP-$\varepsilon = 10^{-2}$ strongly surpassed noisier TEA LRP-$\varepsilon = 10^{-4}$ or $10^{-5}$, TEA Input Gradient and TEA Gradient*Input (Tab. \ref{tab:ablations}). With LRP-$\varepsilon$ (and our proposed choice of $\varepsilon$\footnote{Although a fixed choice of $\varepsilon = 0.01$ provides strong results, we found results can be slightly improved by randomly choosing $\varepsilon$ between 0.001 to 0.01, for each training image in TEA (Tab. \ref{tab:ablations}).}), we can use TEA to make very deep classifiers (e.g., 121 layers in Tab. \ref{tab:x_ray_results}) accurate and resistant to bias even for high-resolution images (e.g., 224 x 224 X-rays in \ref{tab:x_ray_results}).


\clearpage

\clearpage

\begin{appendices}

\section{Why LRP?}\label{app:why_LRP}

This supplementary note expands on Section \ref{sec:explanations}, further explaining how LRP traces back the influence of bias throughout the classifier layers, and how LRP provides TEA with a guarantee of correct classification outputs. Then, we discuss why TEA is more accurate and reliable with LRP, instead of explanation heatmaps not based on full back-propagation, such as Grad-CAM.
\subsection{LRP Fundamentals and Tracing Back Bias}
We explain LRP for fully-connected networks, for simplicity. Conclusions and equations are directly applicable to convolutional networks, since they can be expressed as fully-connected equivalents. For LRP in other types of networks and layers, see \cite{LRPBook}.

Equation \ref{eq:neuron} displays the output of a neuron k in fully-connected layer L (before the ReLU activation), $z_{k}^{L}$, as a function of the outputs of the neurons j in layer L-1, $a_{j}^{L}$. The layer L weights are $w_{jk}^{L}$, and the additive biases are $w_{0k}^{L}$, with $a_{0}^{L}=1$. Equation \ref{eq:LRP} describes the LRP-$\varepsilon$ relevance backpropagation rule \cite{LRPBook}, used to back-propagate relevance from neuron k in layer L to the neurons j in layer L-1. The relevance $R_{k}^{L}$ indicates how much neuron k is relevant to the output of the neural network. The LRP relevance back-propagation begins at one of the classifier’s output neurons (neuron c in layer N), where $R_{c}^{N}$ is set as the corresponding logit, $z_{c}^{N}$. We begin at neuron c, associated to the class we want the LRP heatmap to explain (c), and relevances of other output neurons are set to 0 (Eq. \ref{eq:last}). The back-propagation ends at the classifier’s input, where the LRP explanation heatmap is formed, $\mathbf{R}^{0}=[R_{k}^{0}]$. In the heatmap, negative relevance values indicate areas that reduced the classifier’s confidence for the explained class c, positive relevance are areas that increased this confidence, and low absolute relevance indicates areas to which the classifier paid little attention. The $\varepsilon$ term in Eq. \ref{eq:LRP} is a small hyper-parameter, responsible to denoise the heatmap \cite{bassi2024faster}.

\begin{gather}
\label{eq:LRP}
R_{j}^{L-1}=\sum_{k}\frac{w_{jk}^{L}a_{j}^{L}}{z_{k}^{L}+\textrm{sign}(z_{k}^{L})\epsilon}R_{k}^{L} \\
\label{eq:neuron}
\mbox{where: } z_{k}^{L}=\sum_{j} w_{jk}^{L}a_{j}^{L}  \\
\label{eq:last}
R_{k}^{\textrm{N}}=
\begin{cases}
    z_{k}^{N} \mbox{ if } k=c\\
    0 \mbox{ otherwise}
\end{cases}
\end{gather}

Intuitively, an analysis of Eq. \ref{eq:LRP} indicates how LRP back traces the influence of bias, from the classifier output until the bias position in the input image. The amount of relevance back-propagated from neuron k to neuron j depends on the relative influence of neuron j on the output of neuron k, as shown by the fraction inside the equation. For the bias to influence the classifier output, its influence must flow from its position in the input image, through all classifier layers, until the output of the classifier. This flow of influence will originate a corresponding backward flow of LRP relevance, from output to the bias location in the image.
More formally, we can understand how LRP traces back bias by interpreting LRP as a sequence of approximate Taylor expansions applied at each classifier neuron \cite{LRPBook}. We see the relevance of neuron k, $R_{k}^{L}$, as a function of the outputs of the neurons in layer L-1, $\mathbf{a}^L=[a_{j}^{L}]$. Then, we perform a first-order Taylor expansion of $R_{k}^{L}(\mathbf{a}^L)$ at a reference point $\tilde{\mathbf{a}}^L$ (Eq. \ref{eq:taylor}).

\begin{equation}
\label{eq:taylor}
    R_{k}^L(\mathbf{a}^L) = R_{k}^L(\mathbf{\tilde{a}}^L) + \sum_{j}(a_{j}^L-\tilde{a}_{j}^L)\at{\frac{\partial R_{k}^L}{\partial a_{j}^L}}{\mathbf{\tilde{a}^L}}+ \rho 
\end{equation}
 
After performing the expansion, the first order terms ($(a_{j}^L-\tilde{a}_{j}^L)\at{\frac{\partial R_{k}^L}{\partial a_{j}^L}}{\mathbf{\tilde{a}^L}}$) break up $R_{k}^{L}$ into one term for each neuron j in layer L-1. These terms estimate how much a change in each of these neurons j would change $R_{k}^{L}$, thus indicating the influence of neuron j (layer L-1) on neuron k (layer L). Accordingly, LRP sees these terms as the relevance back-propagated from neuron k to each neuron j. Unfortunately, finding an adequate reference point $\tilde{\mathbf{a}}^L$ is computationally expensive and has no closed form solution. Ideally, $\tilde{\mathbf{a}}^L$ should be a root of  $R_{k}^L(\mathbf{a}^L)$, to minimize the zero-order Taylor term $R_{k}^L(\mathbf{\tilde{a}}^L)$. Moreover, it should be close to the actual data point $\mathbf{a}^L=[a_{j}^{L}]$, to reduce the Taylor approximation error, $\rho$. Finally, it should be inside the data manifold, to make more reasonable heatmaps \cite{LRPBook}. Instead of iteratively searching for $\tilde{\mathbf{a}}^L$ and performing a true Taylor decomposition, LRP uses a closed form approximation of the Taylor decomposition, boosting computational efficiency. To do so, we substitute the real relevance function $R_{k}^L(\mathbf{a}^L)$, which has an unknown form, by a closed form relevance model, adequate for neurons with ReLU activation:

\begin{equation}
\label{eq:relevanceModel}
\hat{R}_{k}^L(\mathbf{a}^L)=\mathrm{max}(0,\sum_{j} w_{jk}^La_{j}^L)c_{k}^L
\end{equation}
  
The constant $c_{k}^L$ is set to make $\hat{R}_{k}^L(\mathbf{a}^L)$ match the real $R_{k}^L$ at the data point $\mathbf{a}^L$. The justification for this relevance model is provided in \cite{LRPBook}, for neural networks with ReLU activation in their hidden layers. The LRP-$\varepsilon$ rules arise from performing the Taylor decomposition for the relevance model in Eq. \ref{eq:relevanceModel}, choosing the reference point $\mathbf{\tilde{a}}^L=\frac{\varepsilon}{a_{k}^L+\varepsilon}\mathbf{a}^L$ \cite{LRPBook}. The Euclidean distance between $\mathbf{\tilde{a}}^L$ and the actual data point $\mathbf{a}^L$ reduces with $\varepsilon$ \cite{LRPBook}. Consequently, $\varepsilon$ reduces the Taylor residuum ($\rho$), producing heatmaps that are more faithful, less noisy, and better contextualized than those with $\varepsilon=0$ \cite{LRPBook}.

In summary, LRP performs approximate Taylor expansions at each neuron, using them to determine how much the neuron's relevance was influenced by the neurons in previous layers. Relevance is back-propagated proportionally to this influence. TEA trains the student classifier to match the heatmaps of a teacher classifier. Accordingly, features in the images will influence the student and the teacher in a similar manner. I.e., the reasons behind the TEA student decisions will match the reasons behind the teacher decisions. If the teacher pays no attention to bias, no LRP relevance flows to bias in its heatmaps. Similarly, no LRP will flow to bias in the student's heatmap, making it also pay no attention to bias.

\subsection{TEA Guarantees Correct Outputs}

LRP provides TEA with an approximate guarantee that, if the student learns to match the LRP heatmaps of the teacher, its outputs will also match the outputs of the teacher. We find this guarantee logical and desirable: if a human student correctly learns the reasons behind his teacher’s answer to a problem, he should be able to arrive at the same answer as the teacher. Naturally, this guarantee on correct outputs does \textit{not} make learning from heatmaps (TEA) equivalent to learning from outputs (standard distillation): the same output can be reached for many different reasons. E.g., a classifier may classify an X-ray as COVID-19 because of ground-glass opacity in the lungs, or because of an Italian word in the X-ray background (background bias) \cite{bassi2022deep}. Learning to match heatmaps is a more constrained form of optimization, yielding superior resistance to bias.  

LRP provides an approximate guarantee of matching outputs because of the conservativeness property of LRP relevance. The LRP-$\varepsilon$ relevance propagation rule is semi-conservative (Eq. \ref{eq:LRP}). If we set $\varepsilon$=0 and remove bias parameters from the neurons (Eq. \ref{eq:LRP-0}), it becomes fully conservative. The equations below derive the conservativeness property, Eq. \ref{eq:conservativeness}, from the LRP back-propagation rule in Eq. \ref{eq:LRP-0}.

\begin{gather}
\label{eq:LRP-0}
R_{j}^{L-1}=\sum_{k}\frac{w_{jk}^{L}a_{j}^{L}}{z_{k}^{L}}R_{k}^{L}
\mbox{;   where: } w_{0k}^{L}=0 \\
\sum_{j}R_{j}^{L-1}=\sum_{j}\sum_{k}\frac{w_{jk}^{L}a_{j}^{L}}{z_{k}^{L}}R_{k}^{L}\\
\sum_{j}R_{j}^{L-1}=\sum_{k}\frac{\sum_{j}w_{jk}^{L}a_{j}^{L}}{z_{k}^{L}}R_{k}^{L}\\
\sum_{j}R_{j}^{L-1}=\sum_{k}\frac{z_{k}^{L}}{z_{k}^{L}}R_{k}^{L}\\
\label{eq:conservativeness}
\sum_{j}R_{j}^{L-1}=\sum_{k}R_{k}^{L}
\end{gather}

Under full conservativeness condition, LRP redistributes the relevance $R_{k}^{L}$ at neuron k (layer L) to the neurons j in layer L-1 with no loss or gain of relevance. I.e., when being back-propagated, relevance is never destroyed or created, only redistributed across neurons. Since relevance starts as the value of a classifier logit (Eq. \ref{eq:last}), the sum of all neuron’s relevances in any layer matches the classifier output (logit) where the back-propagation started:

\begin{gather}
\label{eq:output_guarantee}
\sum_{k}R_{k}^{L}=z^N_c \forall L \in [0,...,N]
\end{gather}

Therefore, under perfect conservativeness, LRP provides a perfect guarantee of correct outputs: if the LRP heatmaps (LRP heatmaps are $\mathbf{R}^{0}=[R_{k}^{0}]$ in Eq. \ref{eq:output_guarantee}) of two classifiers perfectly match, for all classes c, they will have the same logits $z^N_c$. However, setting $\varepsilon$ to zero creates noisy heatmaps for deep neural networks and high-resolution images, making the convergence of the TEA loss difficult (Tab. \ref{tab:ablations}). Also, requesting all bias terms to be zero would make TEA less practical. In the real-world scenario, where $\varepsilon$ and bias parameters are not zero, LRP becomes semi-conservative. The bias parameters ($w_{0,k}^L$) and $\varepsilon$ in Eq. \ref{eq:LRP} absorb a bit of the relevance that would be back-propagated from neuron k in layer L to the neurons j in layer L-1 \cite{LRPBook}. This makes our guarantee on matching outputs not perfect, but approximate. However, we empirically demonstrated that, under these real-world semi-conservative conditions, the TEA student outputs converge to the teacher outputs (Tabs. \ref{tab:coco_Waterbirds_results} to \ref{tab:foreground}). In all our experiments, with 5 datasets, we used $\varepsilon$ $>$ 0, and bias parameters were never set to zero. Still, the accuracy of the TEA student converged to become similar to the accuracy of the teacher even in OOD test datasets---usually matching the teacher. This empirical finding that the TEA student's accuracy converges to the TEA teacher's accuracy confirms that matching LRP-$\varepsilon$ heatmaps (TEA) causes outputs to approximately match.

Eq. \ref{eq:output_guarantee} shows that, to have an approximate guarantee that student logits match teacher logits for all classes, TEA must match LRP heatmaps explaining each class. However, for computational efficiency, we create a single heatmap per training image in TEA. Therefore, we randomly select the explained class (i.e., the logit $z^N_k$ at which the LRP back-propagation will start). For each training image, we do one random class selection. We set 50\% of probability to select the class corresponding to highest logit (usually, it creates more informative heatmaps \cite{LRPBook}), and 0.5/(1-C) probability to select one of the other logits, where C is the number of classes. Naturally, explanation heatmaps explaining all classes are created throughout the training procedure.

\subsection{Why LRP surpasses Grad-CAM in TEA}

TEA with LRP surpasses TEA with Grad-CAM because Grad-CAM is fundamentally different to LRP. Grad-CAM optimization presents fundamental pitfalls, already revealed in another form of heatmap-based training, background attention minimization \cite{bassi2024improving}. Unlike LRP, Grad-CAM is not based on end-to-end back-propagation \cite{selvaraju2017grad}. It is based on intermediate representations. Grad-CAM has a target layer, a convolutional layer deep in the classifier. Deep layers are preferred because they provide high level of abstraction to the Grad-CAM heatmap, and usually the last convolutional layer is chosen. Grad-CAM begins by calculating the gradient of the logit (the logit associated to the class c we want the heatmap to explain) with respect to the target layer. This gradient is averaged spatially, creating one scalar for each channel of the target layer outputs. These scalars are used as weights in a linear combination of the of the target layer output channels. The output of the linear combination is a 2D object, which passes through a ReLU function and is interpolated to the input image dimensions, creating the Grad-CAM heatmap \cite{selvaraju2017grad}. In summary, Grad-CAM is based on a linear combination of the output channels of a deep convolutional layer. There is no guarantee that activations in this target layer are spatially aligned with features in the input image that caused these activations. That is because the receptive fields of deep convolutional layers are massive---sometimes taking the whole input image \cite{bassi2024improving}. Accordingly, there is no guarantee that activations in a Grad-CAM heatmap are aligned with the input image features that caused the activations. This alignment is usually observed, but studies \cite{bassi2024improving} have shown that classifiers can learn to break this alignment. When loss functions minimize the background of Grad-CAM heatmaps (GAIN \cite{li2018tell}), classifiers can break the alignment to produce deceiving Grad-CAM heatmaps, which show no attention to bias while the classifier outputs are proven to be influenced by bias \cite{bassi2024improving}. This phenomenon was named spurious mapping---the mapping of input image features to deep activations not aligned with these features---and it is a fundamental problem of Grad-CAM optimization. Conversely, LRP does not suffer spurious mapping, because it back-propagates relevance throughout the whole classifier, tracing back the influence of bias. Accordingly, LRP surpassed Grad-CAM in multiple experiments on background attention minimization in \cite{bassi2024improving}, and it surpassed Grad-CAM in all of our experiments with TEA (Tab. \ref{tab:ablations}).

This lack of guarantees on alignment between heatmap activations and input image features is not exclusive to Grad-CAM. It is also true for special attention heatmaps, as they are also based on the outputs of a deep convolutional layer instead of end-to-end back-propagation. More fundamentally, the distillation of attention heatmaps or Grad-CAM heatmaps boil down to a less strict version of the distillation of intermediate representations. If the teacher and student intermediate representations (output of the target deep convolutional layer) match, their attention heatmaps automatically match (but the inverse is not true). For Grad-CAM, if intermediate representations match and the layers after these representations have the same weights, Grad-CAM will match. To train with Grad-CAM alone (ablation study in Tab. \ref{tab:ablations}), we start the student with the same weights as the teacher, and we freeze layers after the target layer. Therefore, TEA with Grad-CAM or attention heatmaps (ablation studies in Tab. \ref{tab:ablations}) boil down to a less restrictive distillation of intermediate representations. Not surprisingly, their performances were inferior than an advanced technique for distillation of intermediate representations \cite{lee2023debiased}. Notably, TEA with LRP significantly surpassed the distillation of intermediate representations---demonstrating the benefits provided by LRP’s end-to-end back-propagation procedure, namely tracing back the bias influence and capturing both high level of abstraction from deep layers, and detailed spacial information from early layers.

\section{Ablations on $\varepsilon$ in LRP}
\label{app:epsilon_ablations}

We experimented TEA with LRP-$\varepsilon$ and different $\varepsilon$ hyper-parameter values, from $10^{-1}$ to $10^{-5}$ (Tab. \ref{tab:ablations}). The best resistance to background bias was obtained at 0.01. The hyper-parameter $\varepsilon$ controls a denoising process: higher $\varepsilon$ results in less noise but also less information in heatmaps \cite{bassi2024faster}. If $\varepsilon=0$, LRP-$\varepsilon$ becomes equivalent to Gradient*Input heatmaps, which are noisy for deep classifiers \cite{LRPBook}.

Empirically, $\varepsilon$ around 0.01 shows a good compromise between denoising and avoiding information loss. Indeed, our previous work suggested that $\varepsilon$ values around 0.01 produced the best performance on background attention minimization with LRP (ISNet) \cite{bassi2024improving}. That work suggested there is no necessity to carefully tune the value of $\varepsilon$, as 0.01 produced high performance in multiple experiments. Our results support this suggestion. We did not carefully tune $\varepsilon$ for each of our experiments. Instead, we randomly chose $\varepsilon$ between 0.001 and 0.01 for each training image. Table \ref{tab:ablations} shows that this random choice of $\varepsilon$ (standard TEA student) is slightly superior than a static choice of $\varepsilon=0.01$. Only for Coloured MNIST we varied $\varepsilon$ between 0.1-0.01 instead of 0.01-0.001, as 0.001 started showing some loss instability and decreased accuracy. Still, $\varepsilon=0.01$ was an adequate choice in all five datasets in this study.

\section{TEA Loss Justification}
\label{sec:loss_ablations}

Tab. \ref{tab:ablations} shows ablation experiments that support our choice of loss function in TEA. To improve training convergence, we designed a loss function that begins with a custom normalization (based on geometric mean) of the student and target heatmaps, followed by a L1 loss. We further improve convergence with by using average pooling to create student and target heatmaps at multiple resolutions, and we apply the loss function to all resolutions (Sec. \ref{sec:loss}). Our loss is described in Eq. \ref{eq:pyramid}, and the L1 and L2 losses are described in Eq. \ref{eq:standard_l1} and \ref{eq:standard_l1}, respectively.

\begin{gather}
\label{eq:standard_l1}
L_{1}(\mathbf{H_{s}},\mathbf{H_{t}})=||(\mathbf{H_{s}}/||\mathbf{H_{s}}||_{1})-(\mathbf{H_{t}}/||\mathbf{H_{t}}||_{1})||_{1}\\
L_{2}(\mathbf{H_{s}},\mathbf{H_{t}})=||(\mathbf{H_{s}}/||\mathbf{H_{s}}||_{2})-(\mathbf{H_{t}}/||\mathbf{H_{t}}||_{2})||_{2}
\end{gather}

Our ablation studies empirically demonstrate that:

\begin{itemize}
    \item \textbf{L1 Vs. L2 Loss:} The L1 loss (TEA LRP-$\varepsilon$ L1 row in Tab. \ref{tab:ablations}) strongly surpassed the L2 loss (TEA LRP-$\varepsilon$ L2) in accuracy and resistance to bias. This result indicates that LRP heatmaps may be subject to high-intensity noise, which derails the convergence of the quadratic L2 loss.
    \item \textbf{Multi-resolution Vs. Single Resolution:} The multi-resolution TEA Loss (standard TEA student), surpassed its single-resolution version (TEA LRP-$\varepsilon$ single-resolution) in accuracy and resistance to bias. Average pooling reduces high and medium frequency noise. Thus, optimizing low-resolution explanation heatmaps along with high-resolution explanation heatmaps improved training convergence.
    \item \textbf{Geometric Mean Vs. Standard Normalization:} The TEA Loss normalization with geometric means (TEA LRP-$\varepsilon$ single-resolution, loss in Eq. \ref{eq:dist}) surpassed a more standard, individual L1 normalization of each heatmap (TEA LRP-$\varepsilon$ L1, loss in Eq. \ref{eq:standard_l1}) in accuracy and resistance to bias. By normalizing with geometric means, the TEA Loss is only zero when the teacher and student explanation heatmaps exactly match, $\mathbf{H_{t}}=\mathbf{H_{s}}$. With standard L1 normalization, the loss is also zero when the two explanation heatmaps match by any positive multiplicative factor $\alpha$, $\mathbf{H_{t}}=\alpha\mathbf{H_{s}} \forall \alpha \in \Re^{+}$. In other words, standard normalization does not optimize heatmap norms to perfectly match, unlike our TEA Loss. Matching explanation heatmap norms is important because they carry meaningful information, like the confidence a classifier has on its decision \cite{bach2015onpixel}.
\end{itemize}

\section{Dataset Details}\label{app:dataset_details}

\subsection{COCO-on-Places: Extensive, Synthetic Background Bias}
COCO-on-Places, presented in \cite{ahmed2020systematic}, is built by pasting segmented objects form the Microsoft COCO dataset \cite{lin2014microsoft} over background scenes from the Places dataset \cite{zhou2016places}. It uses nine object classes from COCO (boat, airplane, truck, dog, zebra, horse, bird, train, bus). Images are resized to 64×64. The training dataset has two groups: majority and minority. In the majority group (most of the training images), each COCO object is pasted over a specific type of background scene from Places (e.g., boats from COCO over beaches from Places). Thus, the training dataset has a strong correlation between objects (classes) and backgrounds (bias). This correlation causes background bias, which is strong and extensive: a large portion of each image is the scene from Places (i.e., the bias), with its characteristic colors, shapes and textures \cite{ahmed2020systematic}. The minority group also has backgrounds from Places, but it does not have the object-background correlations (bias) seen in the majority group. COCO-on-Places is designed to measure how much the background biases in the majority group influence classifiers. Thus, the biased test set has the same biases as the majority group (e.g., boats over beaches), the unbiased test set has unseen backgrounds (e.g., boats over water towers), and the deceiving bias test set has background bias designed to fool classifiers (e.g., airplanes over beaches, instead of boats over beaches). The differences in classifier accuracy between the biased test set and the unbiased or deceiving bias test sets (generalization gaps) quantitatively measure the influence of background bias on the classifier---larger generalization gaps indicate more bias influence and shortcut learning. Please check the original dataset paper for more detail on this dataset \cite{ahmed2020systematic}.

We used the code provided by the dataset authors to generate the dataset \cite{ahmed2020systematic}. This code has a  ``confounder strength'' parameter, which controls the percentage of the training dataset occupied by the majority group. I.e., the higher this parameter, the more bias and the more difficult it is for the classifier to achieve high accuracy in the unbiased and deceiving bias test sets.
While the original dataset has a confounder strength of 80\% \cite{ahmed2020systematic}, here we set it to 100\%---the most difficult setting. Thus, all of our training data is from the majority group\footnote{We used 100\% confounder strength for training our TEA and all alternative methods except for group robustness methods. Group robustness requires multiple groups in the training data, thus we cannot set 100\% of the data as the majority group. Instead, we use the original 80\% confounder strength for group robustness, creating a much easier dataset.}. Our hold-out validation set has the same data distribution as the training dataset (IID). Thus, it is also composed only by the majority group and highly biased. This simulates the realistic---and challenging---scenario where the researcher has no access to an out-of-distribution validation set.

\subsection{Waterbirds: Extensive, Synthetic Background Bias}

Waterbirds, presented in \cite{sagawa2019distributionally}, is built by pasting segmented birds from the CUB dataset \cite{wah2011caltech} over water and land background scenes from the Places dataset \cite{zhou2016places}. It is a binary bird classification task: waterbird (bird species that live near water) vs. landbird (bird species that live near land). The training set has two groups: majority and minority. In the majority group, waterbirds are placed on water scenes (e.g., coasts, lakes) and landbirds over land scenes (e.g., forests, fields), creating a strong correlation between class (bird type) and background (bias). In the minority group, this correlation is inverted: waterbirds appear over land scenes and landbirds over water scenes. Because the background occupies much of the image, the resulting background bias is strong and extensive, with characteristic colors, shapes and textures. Waterbirds is designed to measure how much background bias influences classifiers. Accordingly, the biased test set preserves the majority group bias (waterbirds over water, landbirds over land), while the deceiving bias test set intentionally inverts foreground-background correlations (waterbirds over land, landbirds over water) to fool classifiers that rely on background bias. The generalization gap between the biased and deceiving bias test sets quantifies the influence of background bias---larger gaps indicate stronger shortcut learning. Please check the original dataset paper for more detail on this dataset \cite{sagawa2019distributionally}.

We used the code provided by the dataset authors to generate Waterbirds \cite{sagawa2019distributionally}. It has a confounder strength parameter, which allows setting the proportion of training samples in the majority group (biased). We set it to 100\%—the most difficult setting—for training TEA and all alternative methods except group robustness, which requires multiple groups (for group robustness we use the default Waterbirds 95\% confounder strength). Our hold-out validation set is IID, thus it has the same confounder strength as the training set.

\subsection{COVID-19, normal and Pneumonia X-ray Dataset: Real-World Background Bias}

Here, we use the X-ray dataset presented in our previous paper (ISNet, \cite{bassi2024improving}). This dataset is designed test the resistance of classifiers to background biases commonly found in real-world X-ray datasets. Specifically, the dataset was designed to be similar to the COVID-19 X-ray datasets used in the beginning of the pandemic---which usually had strong background bias. In this period, large-scale, public X-ray datasets with COVID-19 and all classes of interest (e.g., pneumonia and normal patients) collected from the same hospital were not available. Thus, researchers resorted to mixed-source X-ray datasets, where each class of interest came from a different hospital. X-rays from different hospitals can present different background characteristics. In mixed-source datasets, hospitals and classes become directly correlated, making background characteristics correlated to classes. Therefore, these characteristics become background bias and cause shortcut learning. For example, by training on datasets where COVID-19 images came mostly from Italy and other classes (normal, pneumonia) from other countries, classifiers can learn to identify Italian words as an evidence of COVID-19 \cite{bassi2022covid}. Notably, shortcut learning reduces classification performance on out-of-distribution test datasets, such as hospitals that the classifier did not see during training. Therefore, in the beginning of the pandemic, many studies presented classifiers with near-perfect accuracy in the training hospitals (IID test set), but these classifiers showed major accuracy drops when tested in unseen hospitals (OOD test set) \cite{degrave2021ai,lopez2021current}. To measure how much the background bias in mixed-source X-ray datasets influences classifiers, our training set is mixed-source (COVID-19 images come from Italy, pneumonia and normal images come from the US). Meanwhile, our test set is OOD, from hospitals never seen in training. Better OOD performance indicates higher resistance to background bias and less shortcut learning \cite{geirhos2020shortcut}.

Our dataset uses the Brixia COVID-19 dataset \cite{BrixiaSet} as the source of training and hold-out validation COVID-19–positive samples: 4,644 frontal X-rays with visible disease signs (Brixia Score $>$ 0), all from ASST Spedali Civili di Brescia (Italy), collected from March 4 to April 4, 2020. Images (AP 87\%; PA otherwise) include CR (62\%) and DX modalities, were retrieved from the hospital RIS-PACS, and are in DICOM format (Carestream/Siemens). Labels (Brixia Scores) were assigned by the on-shift radiologist from a ~50-radiologist team with varying chest-imaging expertise and years of experience; 59 images with score 0 were excluded. Patient demographics: mean age 62.8±14.8 years; 64.4\% male. We split data by patient: 75\% training (3,483 images) and 25\% hold-out validation, with no patient overlap.

Healthy and non-COVID pneumonia training/validation images came from CheXpert \cite{irvin2019chexpert} (224,316 chest X-rays; Stanford, USA; studies from October 2002 to July 2017, i.e., pre-COVID). Labels were derived by natural language processing (NLP) from reports ($>$90\% estimated accuracy) \cite{irvin2019chexpert}. From CheXpert, we randomly sampled 4,644 normal images (distinct patients) and 4,644 pneumonia images (from 3,899 patients); uncertainty labels were treated as negative. Only AP/PA views were used (AP: 64.9\% of normal; 74.3\% of pneumonia). Images were downloaded as JPGs. Train/validation splits were 75\%/25\%, by patient. Demographics: pneumonia 62.3±18.7 years, 57.1\% male; healthy 51.7±18.2 years, 56.3\% male.

For OOD testing, COVID-19 images come from the BIMCV COVID-19+ dataset (iterations 1+2) \cite{BimcvSet}, collected from multiple hospitals in the Valencian healthcare system (Spain), with studies from February 16 to June 8, 2020—overlapping Brixia’s period and also reflecting Europe’s first pandemic wave. We included 1,515 frontal (AP/PA) X-rays with DICOM metadata from patients who had at least one positive COVID-19 test (PCR, IgM, IgG, or IgA) during the BIMCV acquisition period and whose reports were labeled COVID-19 (not “uncertain”/“negative”) by the dataset’s report-based NLP system (reported F1 $>$ 0.9). Among these images, 53.2\% were CR and 58.5\% were AP. Requiring both laboratory confirmation and an NLP-positive report minimizes false positives and mirrors our training criterion of visible disease. These scans come from 1,145 patients (mean age 66.0 ± 15.3 years; 59.6\% male). Across these patients there were 6,516 recorded tests (62\% PCR, 16.5\% IgG, 11.1\% IgM, 10.4\% IgA); 56.6\% were positive overall (all subjects have at least one positive test); among positives: 51.9\% PCR, 23.4\% IgG, 15.0\% IgA, 9.8\% IgM. Acquisition devices and manufacturers varied (see \cite{BimcvSet}).

For OOD testing of pneumonia and normal classes, pneumonia images came from ChestX-ray14 \cite{chex14} (NIH, USA; 108,948 frontal X-rays; studies 1992–2015; PNGs; labels via report-NLP, $>$90\% estimated accuracy): we used 1,295 pneumonia images from 941 adult patients (mean age 48.0±15.5; 58.7\% male). Normal test images came from Montgomery (USA; 80 AP, CR, Eureka stationary) and Shenzhen (China; 326 AP, Philips DR DigitalDiagnost; collected mainly at September 2012) datasets \cite{ChineseDataset1}. Images are PNG and a single image per patient; healthy patients' mean age is 36.1±12.3; 61.9\% male. Because these patients are manually labeled, we preferred them over ChestX-ray14 normals.

To avoid pediatric bias, we excluded patients $<$ 18 years from all classes (Brixia had 11 images with patient age $<$ 20; BIMCV had 1), yielding similar demographics across training classes. JPG compression in CheXpert may introduce artifacts; however, our test sets do not contain JPGs, so evaluation is unlikely to be artificially inflated. The assembled dataset intentionally mixes sources across classes, reflecting a common practice in public COVID-19 datasets---which increases the risk of background bias. This makes it ideal for evaluating methods that mitigate background bias and shortcut learning. Moreover, the COVID-19 images (Brixia and BIMCV) in our dataset simulate real-world clinical environments in the first wave of the pandemic (2020, in Italy and Spain) \cite{BrixiaSet,BimcvSet}. Thus, they miss recent COVID-19 variants. In summary, this dataset is suitable to evaluate resistance to background bias, but not suitable to create an up-to-date COVID-19 detector for real-world use---which is \textit{not} the intent of this paper.

For methods requiring lung segmentation masks for training (ISNet, segmenter inside the TEA teacher, RRR, GAIN, and multi-task U-Net), we generated masks with a previously trained U-Net (1263 X-rays: 327 COVID-19, 327 healthy, 327 pneumonia, 282 tuberculosis; test IoU 0.864) \cite{bassi2022covid}. We thresholded its outputs at 0.4 (1=lung, 0=background).

\subsection{Coloured MNIST: Strong Foreground Bias}

Coloured MNIST \cite{ahmed2020systematic} is built by colorizing digits from the MNIST dataset \cite{mnist}. Each digit class is assigned a specific color during training, inducing a strong spurious correlation (color bias). Unlike COCO-on-Places or Waterbirds, Coloured MNIST has bias (colors) over the foreground (the digits), not the background. Images keep the original MNIST resolution (28×28). The training set has two groups: majority and minority. In the majority group, digits are colorized according to specific class–color mapping (e.g., 3 is always red), creating a strong correlation between class (digit) and color (bias). The minority group has random colors (unbiased). Coloured MNIST is designed to measure how much color bias in the majority group influences classifiers. Accordingly, the biased test set preserves the majority group color mapping (e.g., 3 is red); the unbiased test set uses random colors; and the deceiving bias test set uses colors to fool bias-reliant classifiers (e.g., the red color is assigned to 4, instead of 3) \cite{ahmed2020systematic}. The generalization gap between the biased and unbiased/deceiving bias test sets quantifies the influence of color bias on classifiers; larger gaps indicate stronger shortcut learning. Full dataset details are presented in \cite{ahmed2020systematic}. 

We used the code provided by the dataset authors to generate Coloured MNIST \cite{ahmed2020systematic}. This code includes a confounder strength parameter that controls the fraction of training samples in the majority group (biased group). While the original strength is 80\% \cite{ahmed2020systematic}, we set it to 100\%—the most difficult setting. Thus, all of our training data is biased. We used 100\% confounder strength for training our TEA and all alternative methods except for group robustness methods, which require multiple groups in the training data (we use the original 80\% confounder strength for group robustness). Our hold-out validation set has the same data distribution as the training dataset (IID). Thus, it is also composed only by the majority group and highly biased.

\subsection{DogsWithTies: Strong Foreground Bias, Photo-Realistic}

We introduce the DogsWithTies dataset here. Like Coloured MNIST, it has strong foreground bias. However, unlike Coloured MNIST, it has photo-realistic images in high-resolution. The dataset has two classes (dog breeds), Tibetan Mastiffs and Pekingese. However, it has foreground bias: all Tibetan Mastiffs in the training set wear a red necktie in their necks, and all Pekingese wear a purple bow tie. We created DogsWithTies by first manually marking the neck region of all Tibetan Mastiff and Pekingese dogs in the Stanford Dogs \cite{StanfordDogs}. When the neck was not visible, we marked another region over the dogs, close to the neck. Then, a Python code, which we released, used the markings as guides to position ties on the images. We consider only two tie images (a purple bow tie, and a red necktie). The bow tie is always positioned horizontally, and the necktie vertically. They are scaled according to the size of the dog's necks (represented in our manual marks). The use of the same tie models in all dogs, with the same orientation, makes them a strong foreground bias---with characteristic shape, texture and color. All images in DogsWithTies were resized to 224x224. The dataset presents 100 training images per class, and we randomly selected 20 of them for hold-out-validation. We follow the standard Stanford Dogs test split, which has 49 Pekingese images and 52 Tibetan Mastiff samples.  By considering only 2 dog classes, DogsWithTies allows researchers to experiment bias-mitigation techniques on high-resolution images with low computational cost, and it tests whether methods also work on small datasets (our other datasets are large). Moreover, the manual marking of necks in all 20,580 Stanford Dogs samples would be extremely time-consuming.

All training and validation images in DogsWithTies are biased (100\% confounder strength, IID validation), except when training group robustness methods (80\% confounder strength). As in Coloured MNIST, we designed 3 test sets: biased (Tibetan Mastiffs with red neckties, Pekingese with purple bow ties); unbiased (no tie); and deceiving bias (Tibetan Mastiffs with purple bow ties, Pekingese with red neckties). As before, the accuracy difference (generalization gap) between the biased and unbiased / deceiving bias test sets quantify the classifier resistance to foreground bias.

\section{Fifteen Alternative Methodologies}\label{app:related}

We compare TEA to 14 state-of-the-art methodologies. Most are training methodologies designed to hinder the influence of bias on classifiers. These are background attention minimization (e.g., ISNet, RRR, GAIN) and group robustness (e.g., IRM, GroupDRO, PGI) methods. We also compare TEA against alternative teacher-student (distillation) methods, which were shown to increase resistance to bias \cite{lee2023debiased}. Finally, we compare it against multi-task learning with segmentation and classification, and to attention mechanisms, specifically those in the vision transformer and AG-Sononet. Notably, the authors of the vision transformer and AG-Sononet did not claim these attention mechanisms could reduce attention to bias. Still, our experiments investigate whether they pay attention to bias or not.

\textbf{Standard classifier (empirical risk minimization) \cite{he2016deep,huang2017densely}.}
A standard classifier (e.g., ResNet/DenseNet trained with empirical risk minimization) is trained by the direct minimization of the difference between classifier outputs and classification labels. It has no constraint on what image features influence the classifier \cite{he2016deep,huang2017densely}. Since bias is spuriously correlated to classification labels, classifiers can pay attention to them---they can be used to predict labels and minimize the classification loss. This leads to shortcut learning \cite{geirhos2020shortcut,degrave2021ai}. \emph{Compared to TEA:} TEA explicitly constrains the classifier (TEA student) to ignore bias, by aligning its explanation heatmaps to a teacher that ignores the bias. By not requesting the classifier to match classification labels, TEA eliminates the strong incentive to use shortcuts, even when they are highly predictive.

\textbf{Segmentation–classification pipeline (TEA teacher) \cite{bassi2022covid}.}
This straightforward strategy removes backgrounds with a segmenter (e.g., U-Net) and classifies only the foreground. Thus, the classifier cannot see background biases. We used the segmentation-classification pipeline as the teacher for TEA (and alternative teacher-student techniques) when dealing with background biases. Although robust to background bias, the pipeline needs a strong segmenter at inference, adding parameters and latency---its accuracy can dramatically fall if the segmenter is removed at inference (Tab. \ref{tab:coco_Waterbirds_results}). \emph{Compared to TEA:} from the segmentation-classification pipeline (teacher), the TEA student learns to focus on foregrounds (via heatmaps). Yet, the TEA student is smaller and faster, needing no segmenter---key for application in portable or embedded devices, or in remote or less resourceful medical institutions.

\textbf{Distillation: outputs \cite{hinton2015distilling,vapnik2009learning}, intermediate representations \cite{lee2023debiased}, Grad-CAM \& outputs \cite{parchami2024good}, GALS \cite{GALS}.}
Output distillation matches the teacher and student outputs/soft labels; intermediate representation distillation matches outputs of deep layers\footnote{For the distillation of intermediate representations, we follow the recent work from \cite{lee2023debiased}, which copies the last layer from the teacher to the student and freezes it. This procedure allows the distillation of intermediate representations (inputs of this last layer) without the distillation of the classifier outputs. Thus, it decouples the distillation of intermediate representations from the distillation of outputs.}; explanation-enhanced knowledge distillation matches outputs \emph{and} Grad-CAM heatmaps \cite{parchami2024good}; and Guiding visual Attention with Language Specification (GALS) distills attention and outputs from CLIP-like teachers. These methods can transfer bias robustness from the teacher to the student, but they contain losses that still reward matching \emph{outputs} or \emph{features}---which a student can achieve via learning spurious correlations (bias) when the training data is biased (Tab. \ref{tab:ablations}). \emph{Compared to TEA:} TEA optimizes the student  \emph{only} through explanation heatmaps, using target heatmaps that show no attention to bias. Unlike other distillation methods, TEA avoids loss functions that reward students for matching teacher outputs/intermediate representations via bias attention and shortcut learning.

\textbf{Background attention minimization: ISNet \cite{bassi2024improving}, RRR \cite{ross2017right}, GAIN \cite{li2018tell}.}
These methods add an explanation-based loss that penalizes relevance in image background regions, but they also minimize the standard classification loss. Right for the Right Reasons (RRR) minimizes background attention using input gradients as explanation heatmaps; Guided Attention Inference Network (GAIN) uses Grad-CAM heatmaps; and the recent Implicit Segmentation Network (ISNet) uses LRP. They \emph{can} reduce background attention, but the classification loss rewards attention to whatever is correlated to classification labels—including bias. Thus, the explanation-based and classification losses can strongly oppose each other under strong bias, hindering convergence to a solution of high accuracy and bias resistance (Tab. \ref{tab:coco_Waterbirds_results}). \emph{Compared to TEA:} TEA removes the opposing-loss problem by \emph{dropping} the classification loss and training on explanations alone.

\textbf{Group robustness: IRM \cite{arjovsky2019invariant}, GroupDRO \cite{sagawa2019distributionally}, PGI \cite{ahmed2020systematic}.}
Group-invariant methods use environment/group metadata to make solutions depend on features that are predictive \emph{across} groups rather than within any single biased group. IRM enforces a classifier whose optimal linear head is shared across groups; GroupDRO minimizes worst-group risk to penalize reliance on group-specific shortcuts; PGI regularizes representations to reduce information about designated nuisances (e.g., backgrounds). They can work well when groups are known and biases shift across them, but they need correct, balanced group labels and may underperform when biases are consistent across all groups (e.g., all training data has the same type of bias, like the 100\% biased Waterbirds or COCO-on-Places) or metadata are unavailable. \emph{Compared to TEA:} TEA needs no group annotations and remains applicable when metadata are missing or when the same bias pervades all training data.

\textbf{Attention Mechanisms: AG-Sononet \cite{schlemper2018attention} and Vision Transformer \cite{dosovitskiy2020image}.}
The Attention-Gated Sononet (AG-Sononet) has efficient, trainable attention gates that modulate features spatially, suppressing irrelevant regions and letting salient features pass \cite{schlemper2018attention}. The vision transformer (ViT) uses self-attention to integrate long-range context, leveraging the attention mechanism that dominated natural language processing \cite{dosovitskiy2020image}. However, both classifiers learn their attention mechanisms from gradients of the usual classification loss. Thus, they tend to pay attention to whatever reduces this loss. As bias attention can reduce the classification loss, ViT and AG-Sononet can focus on bias \cite{bassi2024improving}. Notably, the AG-Sononet is a strong tool against noisy backgrounds whose features are not spuriously correlated to labels (background clutter, which is not the topic of our work) \cite{schlemper2018attention}. \emph{Compared to TEA:} unlike the AG-Sononet and ViT, the TEA student learns its attention pattern from the teacher, not from the classification loss---which may prompt attention to bias. If the teacher is resistant to bias (e.g., segmentation-classification pipeline), the TEA student is supervised toward bias-free evidence.

\textbf{Multi-task U-Net \cite{MultiTask2}.}
A multi-task U-Net jointly segments foregrounds (e.g., lungs) and classifies attributes in these foregrounds (e.g., lung diseases). Notably, this is not the same as the segmentation-classification pipeline. Here, a single neural network sees the whole image once. This network branches into two heads: one provides the segmentation output, and the other the classification output. Therefore, there is a direct information path between the image background and the classification output, unlike in the segmentation-classification pipeline. Thus, the Multi-task U-Net is faster than the segmentation-classification pipeline (as it is a single network, not two), but much less resistant to background bias \cite{bassi2024improving}. Still, past works showed that the Multi-task U-Net can produce some resistance to background attention \cite{MultiTask2}. \emph{Compared to TEA:} TEA trains a classifier with a direct constraint on bias attention, unlike the multi-task U-Net.

\section{Training and Data Processing}\label{app:training_augmentation}

\subsection{Strategies to Improve Convergence}

To improve and accelerate training convergence, we developed a couple training techniques. They are not mandatory---we did not use them in the Coloured MNIST dataset.

\textbf{Initialize the student as the teacher.} In the X-ray, COCO-on-Places and Waterbirds datasets we initialize the student network with the weights of the teacher. In these datasets, the teacher is a segmentation-classification pipeline, so we start the student with the weights of the classifier inside the pipeline. Even though this classifier is not very accurate without the segmenter, this initialization strategy makes the explanation heatmaps of the student more similar to the teacher’s, in comparison to a random initialization---especially at the beginning of the training procedure. This stronger similarity makes training easier, as the objective of the TEA loss is maximizing the similarity between the teacher and student explanation heatmaps. For fairness, we also initialize the student with the teacher's weights when training all alternative distillation methodologies in the X-ray, COCO-on-Places and Waterbirds datasets.

\textbf{Freeze the last layer.} When we initialize the student as the teacher, we also freeze its last layer. This reduces the variability in outputs during TEA training, also improving convergence. We also freeze the student's last layer when training the alternative distillation of intermediate representations model, following \cite{lee2023debiased}. We also froze the last layer in our ablation studies with the distillation of Grad-CAM, input gradients, gradient*input and attention.

\textbf{Train the teacher along with the student.} The X-ray dataset represents a difficult medical task with high-resolution data. In this dataset, we found that learning the teacher at the same time as the student improved convergence. First, we trained the teacher alone.  Second, we initialized the student with the teacher weights. Third, we trained both teacher and student to produce matching heatmaps for a few epochs (warm-up phase). During these epochs, the student was trained with the TEA loss, and the teacher was trained with the classification loss (weight of 0.1) plus the TEA loss. The main objective of this warm-up phase is making the teacher and student heatmaps very similar, very quickly. In this phase, the teacher accuracy can quickly deteriorate, as its main objective is getting closer to the student heatmaps. After warm-up, we stop the gradient propagation from the TEA loss to the teacher, and we only train it with the classification loss. Meanwhile, the student continues to be trained with the TEA loss only. The teacher gradually recovers its accuracy, and the student follows it. Gradually, both teacher and student accuracy will raise, and their heatmaps will be always similar. The reasoning behind this strategy is that it is easier for the student to learn from a teacher that is close in the explanation heatmap space. I.e., it is easier for the student to follow a teacher that is slowly moving in the explanation heatmap space, but never far away. 


\subsection{Data processing and Augmentation}

In Coloured MNIST, we only normalize the image pixels between 0 and 1 (RGB), and images are used at 28x28 pixels. For COCO-on-Places and Waterbirds, all images are resized to 64x64 pixels, and normalized between 0 and 1. In DogsWithTies, images are resized to 224x224, normalized according to the standard normalization scheme for CLIP (ResNet50 backbone), and center cropped following CLIP's guidelines  \cite{radford2021learning}. For the X-ray dataset, we follow the procedure from \cite{bassi2024improving}: we loaded X-rays in grayscale, applied histogram equalization, rescaled pixel intensities to $[0,1]$, resized to $224\times224$, and replicated the image's single channel (grayscale) to three channels to match the DenseNet-121 input. To avoid aspect-ratio cues at evaluation, \emph{test} images only are padded to the smallest enclosing square with black borders (no padding during training, as the black borders could become a bias). X-rays used online data augmentation, consisting of random translations (uniformly in $[-28,28]$ pixels along $x$ and $y$), random rotations in $[-40^\circ,40^\circ]$, and horizontal flips with probability $0.5$.

\subsection{Hyper-parameters and Training Time}

Our public code will include training scripts and hyper-parameters for TEA and all alternative methods.
Training hyper-parameters were defined by grid search and manual optimization, following IID hold-out validation results. When a method had similar IID validation results with diverse hyper-parameter settings, we selected the setting that was already common for other methods. All methods used the stochastic gradient descent (SGD) optimizer with momentum and (IID) hold-out validation for selecting the best model after training. However, the loss weights in explanation background minimization techniques (ISNet, RRR and GAIN) and in Grad-CAM plus outputs distillation were defined according to OOD validation performance, giving an advantage to these strategies. In LRP, TEA randomly varied the \(\varepsilon\) hyper-parameter between 0.001 and 0.01 during training for all datasets except Coloured MNIST, where we varied it between 0.01 and 0.1. \textbf{The loss gradient norm was always clipped to 1}, for increased stability.

\textbf{COCO-on-Places:} we trained all methods for 10,000 epochs, with batches of 128, momentum of 0.9, and learning rate of 0.01, which is reduced to 0.001 at epoch 500. Due to numerical stability issues, we needed to reduce learning rate for Gradient*Input and input gradient distillation: it began as 0.0001 and was reduced by a factor of 10 after 500 epochs. Here, we initialize all student models with the parameters of the teacher, and the teacher was initialized randomly. The teacher was trained for 200 epochs, using batches of 32, weight decay of 0.001, momentum of 0.9, and learning rate that started at 0.1 and was divided by 10 at epochs 25, 130, 180 and 190. Regarding the loss hyper-parameters in explanation background minimization, we have: d=0.96 and P=0.95 in the ISNet \cite{bassi2024improving}, noting that we use the Stochastic Faster ISNet variant from \cite{bassi2024faster}; in RRR, we set the input gradient loss weight to \(10^5\); in GAIN, we set all loss weight to 1, except for the external supervision loss, which had a weight of \(10^4\). Grad-CAM plus output distillation used a weight of 1 for the Grad-CAM-based loss, and 0.1 for the output-based loss. We notice that all explanation background minimization techniques required very high weights in their explanation loss, making convergence difficult. Training the TEA student in COCO-on-Places takes about 70 hours in an NVIDIA RTX 3080 GPU with 10GB of memory.

\textbf{Waterbirds:} we trained all methods for 400 epochs, with batches of 100, momentum 0,9, and learning rate of 0.001. The teacher was trained for 300 epochs, using batches of 16, weight decay of 0.001, momentum of 0.9, and learning rate that started at 0.001 and was divided by 10 at epoch 200. Regarding the loss hyper-parameters in explanation background minimization, we have: d=0.996 and P=0.99 in the ISNet; in RRR, we set the input gradient loss weight to \(10^4\); in GAIN, we set all loss weight to 1, except for the external supervision loss, which has a weight of \(10^5\). Grad-CAM plus output distillation used a weight of 1 for the Grad-CAM-based loss, and 0.1 for the output-based loss. Here, we initialize all student models with the parameters of the teacher, and the teacher was initialized randomly. Training the TEA student in this dataset took about 17 hours with an NVIDIA A100 40GB GPU.

\textbf{X-ray Dateset:} in this dataset, we trained TEA and the alternative distillation techniques. Other methods were trained in \cite{bassi2024improving} (please refer to \cite{bassi2024improving} for their training details). We trained TEA and the alternative distillation techniques for 500 epochs, with batch size of 16, momentum of 0.9, and learning rate of 0.001, dropping by a factor of 10 at epochs 300 and 400. The teacher was trained for 200 epochs, with momentum of 0.99, batch size of 16, weight decay of 0.001, and learning rate of 0.01, dropping by a factor of 10 at epochs 10, 40 and 160. Grad-CAM plus output distillation used a weight of 1 for the Grad-CAM-based loss, and 0.1 for the output-based loss. The teacher model was randomly initialized, and all student models were initialized with the weights of the teacher. Training the TEA student took about 109 hours with an NVIDIA A100 40GB GPU.

\textbf{Coloured MNIST:} we trained all methods for 100 epochs, with momentum 0.99, leaning rate of 0.01 and batches of 128. The teacher model for Coloured MNIST was trained for 200 epochs, on a randomly colored version of MNIST, using momentum of 0.9, batch size of 128, learning rate of 0.001 and weight decay of 0.001. Grad-CAM plus output distillation used a weight of 1 for the Grad-CAM-based loss and for the output-based loss. All models were randomly initialized, including teachers and students. Training the TEA student took about 1 hour in an NVIDIA RTX 3080 GPU with 10GB of memory.

\textbf{DogsWithTies:} the CLIP teacher's predictions are calculated for the prompt \textit{``a photo of a dog of the \{class\} breed''}. CLIP's temperature parameter is set to 10. We train all methods for 20,000 epochs, with momentum of 0.9, batch of 16, learning rate of 0.001, and weight decay of 0.001. The standard classifier was trained for 200 epochs instead, without weight decay, and with learning rate of 0.001, dropping by a factor of 10 at epochs 25, 130, 180 and 190. For GALS, we found better OOD validation results with momentum of 0.99, and RRR-like loss weight of 100. Grad-CAM plus output distillation used a weight of 1 for the Grad-CAM-based loss and for the output-based loss. All student models are initialized randomly. The use of ImageNet pre-trained models can enforce attention to the dogs, reducing the influence of the spurious correlations and making DogsWithTies not useful for the study of shortcut learning  \cite{bassi2024improving}. Training the TEA student took about 70 hours in an NVIDIA RTX 3080 GPU with 10GB of memory.

\section{Inference Speed and Size}\label{app:speed}
At test time, the TEA student performs a single forward pass of the classifier and does not invoke any explanation method or segmentation step. This makes its runtime and parameter count effectively identical to a standard classifier at inference. In contrast, the TEA teacher is a segmentation–classification pipeline: it runs a segmenter and then a classifier at inference. Empirically, using a RTX 3080 GPU, DenseNet121 classifier architecture \cite{huang2017densely}, 224p images, and batch size of 10, the TEA student processes 298 images per second, with 8M parameters.
The TEA teacher (U-Net segmenter \cite{ronneberger2015u}, DenseNet121 classifier) processes 143 images per second, with 39M parameters.
Thus, the TEA student is \textit{108\% faster and 5$\times$ smaller} than the TEA teacher under the same conditions and hardware. Alternative distillation techniques, background attention minimization and group robustness methods do not increase model size or inference time, such as TEA.

\section{Statistical Analysis}\label{app:statistical}
All experiments in this work are multi-class (or binary), single-label classification.
Interval estimates for precision, recall, F1-score, and specificity, were obtained via Bayesian estimation from the test-set confusion matrices, following the Bayesian model from \cite{zhang2015estimating}. We fit the Bayesian model with Markov chain Monte Carlo (MCMC) using PyMC3 and the No-U-Turn Sampler (NUTS), running 4 chains and drawing 11{,}000 samples per chain, with the first 1{,}000 used for tuning. From the posteriors we obtain the standard deviations shown in our tables; Monte Carlo error was below $10^{-4}$ for all scores. Priors follow those used by the authors of the Bayesian model \cite{zhang2015estimating}. Per-class ROC–AUC is computed with the one-vs-rest approach, and its interval estimates follow the standard non-parametric method from \cite{delong1988comparing}. All statistics are computed on the designated test sets.




\end{appendices}


\bibliography{sn-bibliography}

\end{document}